\documentclass[11pt,a4paper]{article}
\usepackage{times,latexsym}
\usepackage{url}
\usepackage{authblk} 
\usepackage[T1]{fontenc}
\usepackage{caption}
\usepackage{float}
\usepackage{subcaption}

\usepackage[acceptedWithA]{tacl2021v1}

\makeatletter

\newcommand\email[2][]%
   {\newaffiltrue\let\AB@blk@and\AB@pand
      \if\relax#1\relax\def\AB@note{\AB@thenote}\else\def\AB@note{\relax}%
        \setcounter{Maxaffil}{0}\fi
      \begingroup
        \let\protect\@unexpandable@protect
        \def\thanks{\protect\thanks}\def\footnote{\protect\footnote}%
        \@temptokena=\expandafter{\AB@authors}%
        {\def\\{\protect\\\protect\Affilfont}\xdef\AB@temp{#2}}%
         \xdef\AB@authors{\the\@temptokena\AB@las\AB@au@str
         \protect\\[\affilsep]\protect\Affilfont\AB@temp}%
         \gdef\AB@las{}\gdef\AB@au@str{}%
        {\def\\{, \ignorespaces}\xdef\AB@temp{#2}}%
        \@temptokena=\expandafter{\AB@affillist}%
        \xdef\AB@affillist{\the\@temptokena \AB@affilsep
          \AB@affilnote{}\protect\Affilfont\AB@temp}%
      \endgroup
       \let\AB@affilsep\AB@affilsepx
}
\makeatother

\usepackage{xspace,mfirstuc,tabulary}

\newif\iftaclinstructions
\taclinstructionsfalse 
\iftaclinstructions
\renewcommand{\confidential}{}
\renewcommand{\anonsubtext}{(No author info supplied here, for consistency with
TACL-submission anonymization requirements)}
\newcommand{\instr}
\fi

\iftaclpubformat 

\else

\fi

\newcommand\github{\raisebox{-2pt}{\includegraphics[width=1.5em]{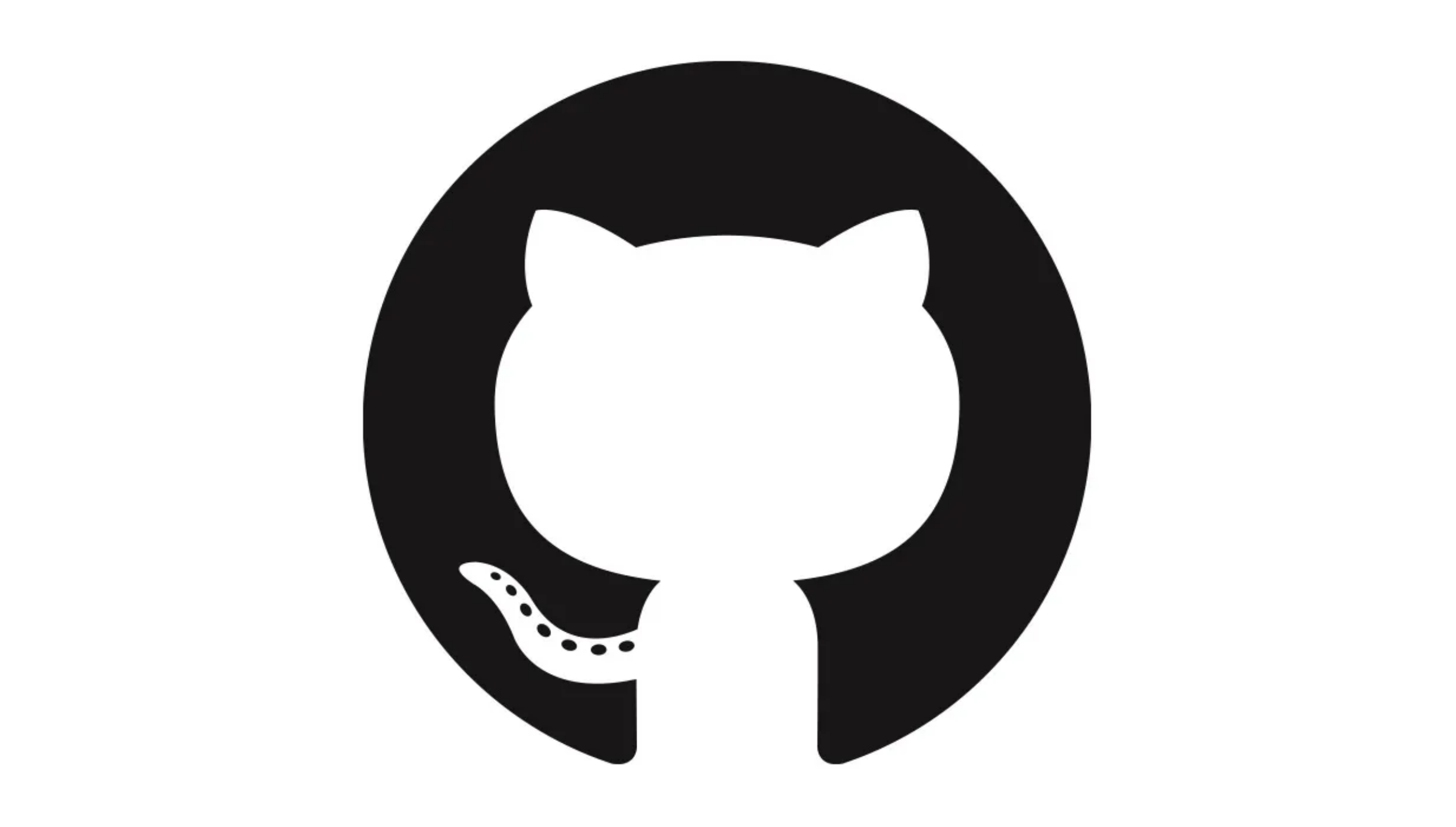}}}
\newcommand\huggingface{\raisebox{-2pt}{\includegraphics[width=.9em]{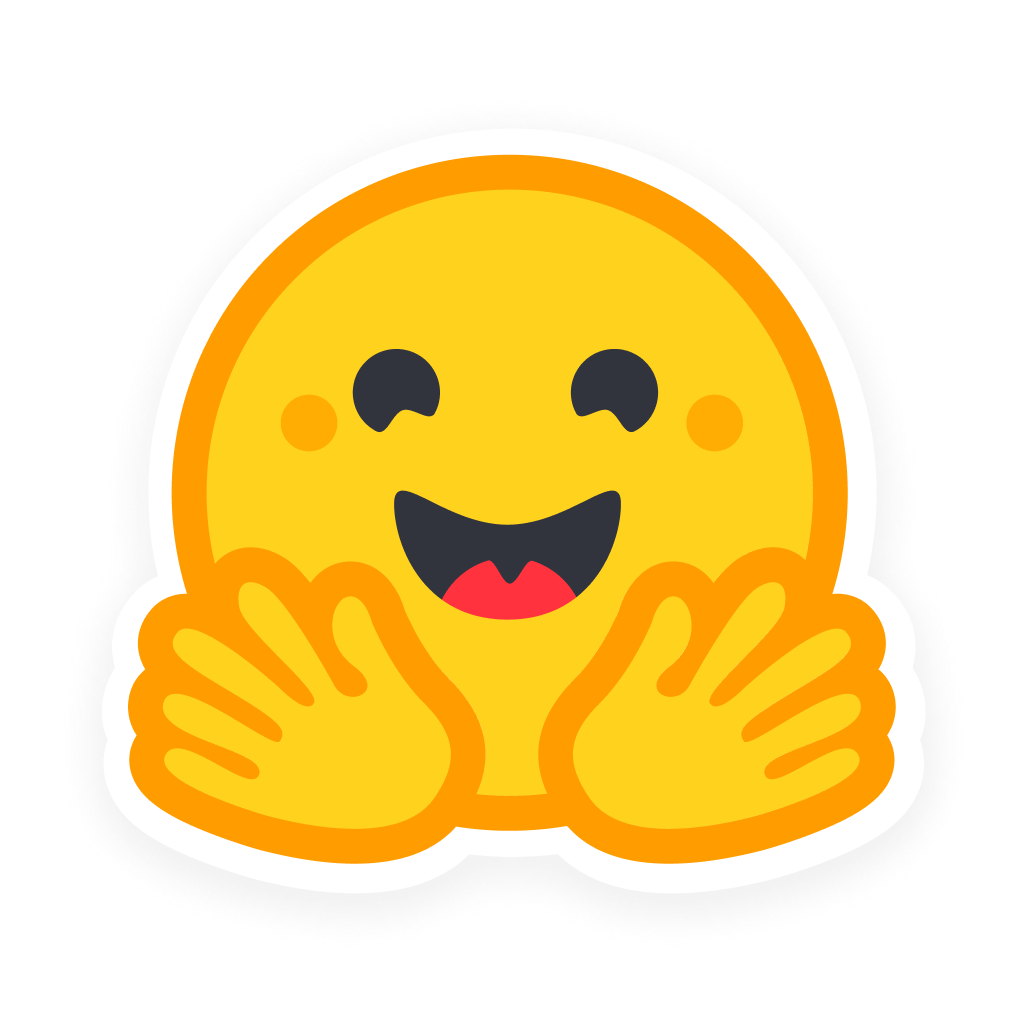}}}
\newcommand\telaviv{\raisebox{.7ex}{$\tau$}\hspace{0.2em}}
\newcommand\mila{\raisebox{.7ex}{$\omega$}\hspace{0.2em}}
\usepackage{graphicx}
\usepackage{booktabs}
\usepackage{multirow} 
\usepackage{amsmath}
\usepackage{xcolor}
\usepackage{enumitem}
\usepackage{longtable}
\usepackage{hyperref}

\definecolor{lightgreen}{HTML}{64b941}
\definecolor{lavender}{HTML}{b4a7d6}
\definecolor{peach}{HTML}{f4b183}
\colorlet{darkpeach}{peach!90!black}
\definecolor{pink}{HTML}{f669a1}
\colorlet{darkpink}{pink!90!black}

\definecolor{learned}{HTML}{98CA82}
\definecolor{forgotten}{HTML}{F09696}

\newcommand{\method}{\texttt{LMEnt}}

\newcommand{\pushpin}{\includegraphics[height=9pt]{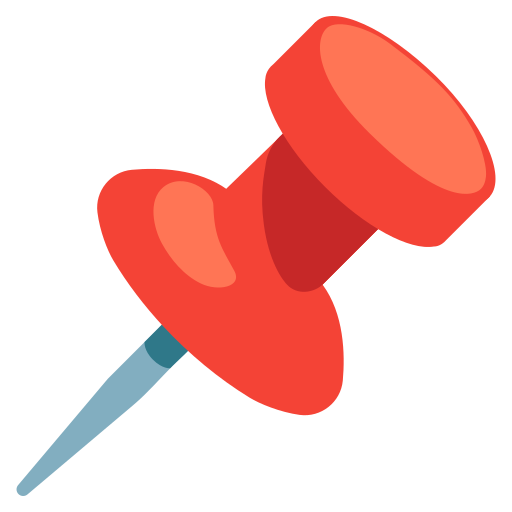}}

\newcommand{\leftarrowemoji}{\includegraphics[height=9pt]{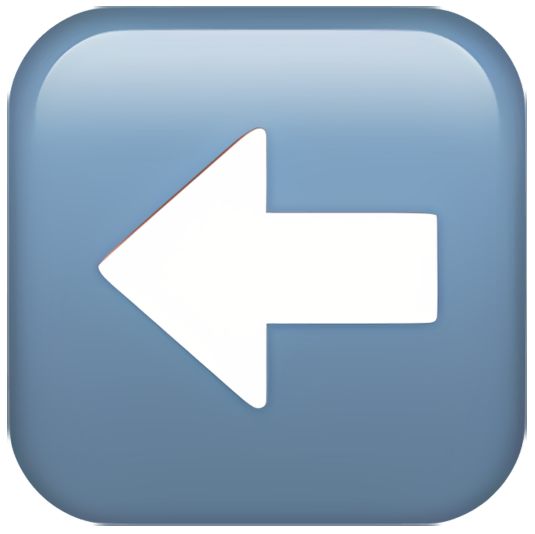}}

\newcommand{\link}{\includegraphics[height=9pt]{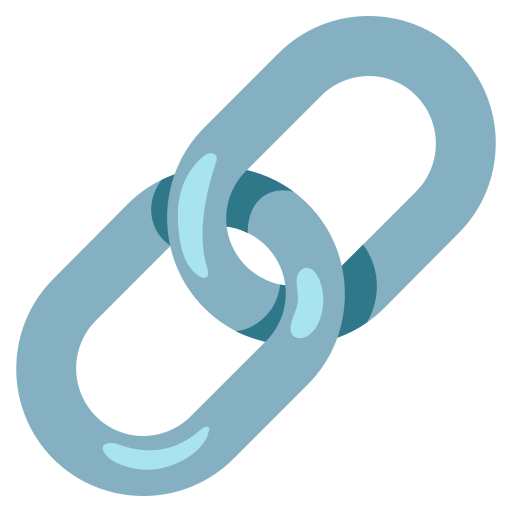}}

\usepackage[most]{tcolorbox}
\usepackage{xcolor}

\definecolor{darkgraybox}{HTML}{4A4A4A}
\definecolor{lightgraybox}{HTML}{EDEDED}

\usepackage{cleveref}
\crefname{figure}{Fig.}{Figs.}
\Crefname{figure}{Fig.}{Figs.}
\crefname{table}{Tab.}{Tabs.}
\Crefname{table}{Tab.}{Tabs.}
\crefformat{subsection}{#2§#1#3}
\Crefformat{subsection}{#2§#1#3}

\crefformat{section}{#2§#1#3}
\Crefformat{section}{#2§#1#3}

\usepackage{xspace}
\usepackage{todonotes}
\usepackage[normalem]{ulem}

\makeatletter
\newcommand*\iftodonotes{\if@todonotes@disabled\expandafter\@secondoftwo\else\expandafter\@firstoftwo\fi}  
\makeatother

\title{\method{}: A Suite for Analyzing Knowledge in Language Models \\ from Pretraining Data to Representations}

\author{
  {\bf Daniela Gottesman}\telaviv \quad
  {\bf Alon Gilaie-Dotan}\telaviv \quad
  {\bf Ido Cohen}\telaviv \quad
  {\bf Yoav Gur-Arieh}\telaviv\\[0.5em] 
  {\bf Marius Mosbach}\mila \quad
  {\bf Ori Yoran}\telaviv \quad
  {\bf Mor Geva}\telaviv
}
\affil{\telaviv Tel Aviv University \hspace{1.5em} \mila Mila -- Quebec AI Institute \& McGill University}

\email{\small{\texttt{gottesman3@mail.tau.ac.il}}}

\date{}

\begin{document}
\maketitle

\begin{abstract}
Language models (LMs) increasingly drive real-world applications that require world knowledge.
However, the internal processes through which models turn data into representations of knowledge and beliefs about the world are poorly understood. To facilitate such studies, we present \method{}, a suite including (1) a knowledge-rich pretraining corpus, fully annotated with entity mentions based on Wikipedia, (2) an entity-based retrieval method over pretraining data that outperforms existing tools by as much as 80.4\%, and (3) 12 pretrained LMs with up to 1B parameters and 4K intermediate checkpoints, with comparable performance to popular open-source models on knowledge tasks. 
Together, these resources provide a controlled environment for analyzing connections between entity mentions in pretraining data and downstream performance. We show the utility of \method{} by studying knowledge acquisition over training, finding that entity co-occurrence and mention forms—which are difficult to study with existing tools—affect learning trends.
Moreover, as LMs form stronger associations between entities, their facts are harder to edit in-context, whereas inconsistencies in model predictions over training are indicative of editing success.
We release \method{} to support studies of knowledge in LMs, including knowledge representations, plasticity, editing, attribution, hallucinations, and learning dynamics.

\vspace{-0.2cm} 
\begin{center}
  \huggingface{} \href{https://huggingface.co/collections/dhgottesman/lment-68a9dd370e1f746cacd8ce58}{\texttt{huggingface.co/LMEnt}} \\
  \github{} \href{https://github.com/dhgottesman/LMEnt}{\texttt{github.com/LMEnt}}
\end{center}

\end{abstract}

\section{Introduction}

Language models (LMs) capture substantial amounts of knowledge and beliefs about the world from their training data \citep[]{llm-knowledgebase, roberts-etal-2020-much, review-llm-knowledge, llm-agent}. 
A fundamental, yet underexplored, question is how such knowledge representations are formed and shaped during pretraining. Namely, what is the interplay between data composition, training dynamics, and the internal knowledge mechanisms in LMs?
An understanding of these processes could provide better control over the model's knowledge, potentially improving its factuality and reasoning. 

\begin{figure*}[t]
\setlength\belowcaptionskip{-9pt}
    \centering
    \includegraphics[width=\linewidth]{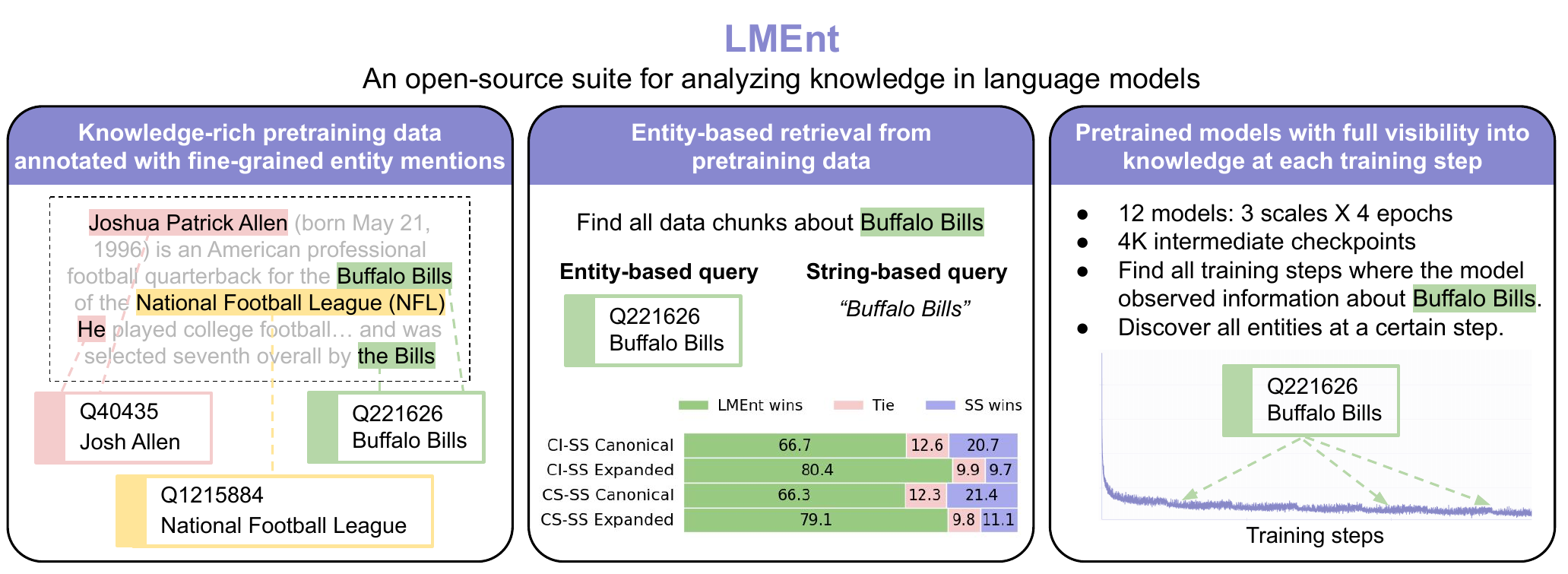}
    \caption{The \method{} suite is composed of three components: (left) fine-grained entity mentions for every document in the pretraining corpus, (middle) an index that retrieves by the entity QID and outperforms string-based retrieval methods, (right) 12 models trained on 1, 2, 4, and 6 epochs where each step can be mapped to the entities it mentions, and each entity can be traced to the steps that introduce it.
    }
    \label{fig:intro}
\end{figure*}

\begin{figure*}[t]
    \centering
    \includegraphics[width=\linewidth]{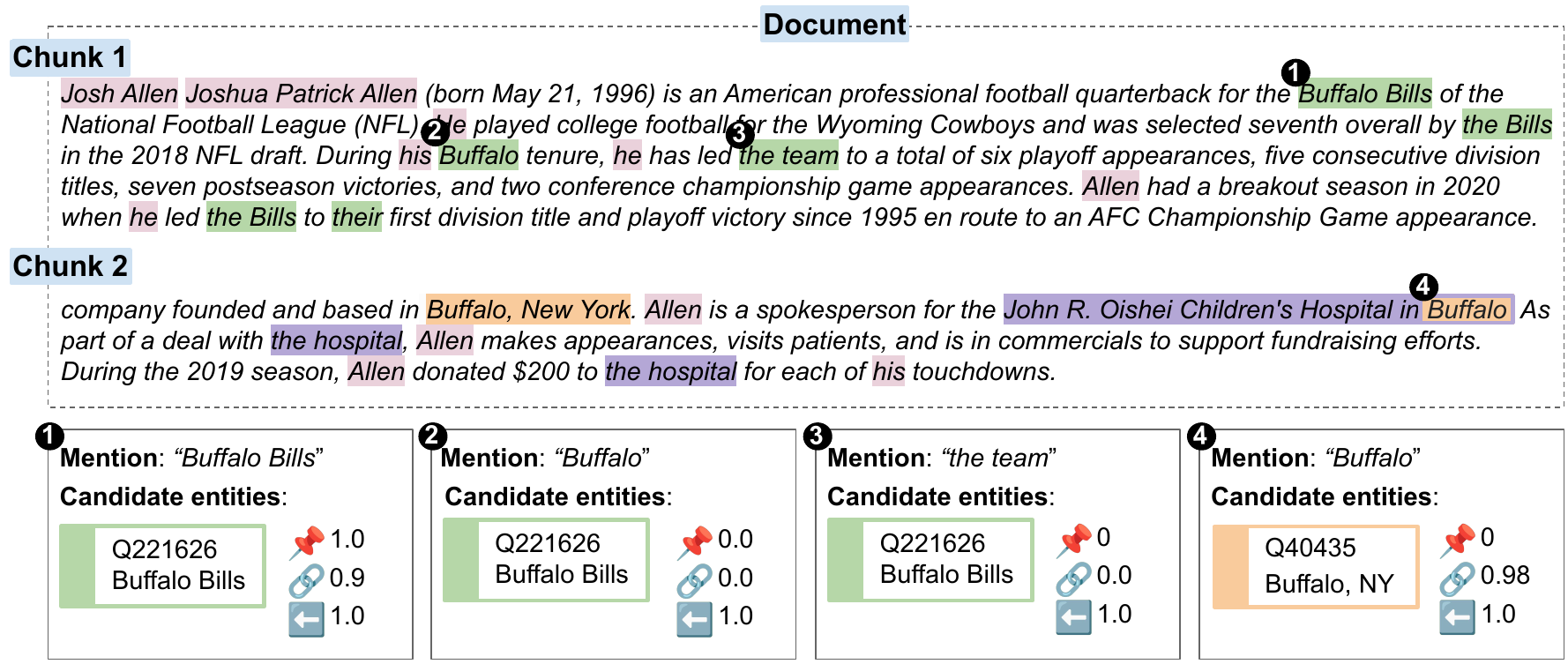}
    \caption{The document about \textcolor{darkpink}{\textit{``Josh Allen''}} is split into two chunks that are processed independently during pretraining. In Chunk 1, \textcolor{lightgreen}{\texttt{Q221626}} is explicitly mentioned in Mention (1), found by hyperlinks and entity linking, and implicitly mentioned in Mention (3), found by coreference resolution. Although \textcolor{lightgreen}{\texttt{Q221626}} (the football team) and \textcolor{darkpeach}{\texttt{Q40435}} (the city of Buffalo) share the surface form ``\textit{Buffalo},'' \method{} correctly disambiguates them: Mention (2) is resolved to \textcolor{lightgreen}{\texttt{Q221626}} with 1.0 confidence through coreference resolution, whereas Mention (4) is linked to \textcolor{darkpeach}{\texttt{Q40435}} with 0.98 confidence using entity linking. 
}
    \label{fig:LMEnt_scoring}
\end{figure*}

A key prerequisite for studying this question is the ability to locate exactly where specific knowledge appears in the pretraining data. However, since such metadata is rarely available for pretraining corpora, many works that study knowledge acquisition have used synthetic data \citep[]{kassner-etal-2020-pretrained, dmtriplets, chang-etal-2024-how, physicsofllm, zucchet2025how, ou-etal-2025-llms}, or post-hoc string-based search tools \citep{wimbd, infinigram, olmotrace} which lack robustness against variability in phrasing of semantically equivalent information. For example, when retrieving information about the American football team the \textit{``Buffalo Bills''} (\Cref{fig:LMEnt_scoring}), these tools are less likely to retrieve documents that refer to the team as \textit{``Buffalo''}, \textit{``The Bills''}, or even \textit{``the team''}, and suffer from noisy retrieval when queries are expanded with name aliases, as we observe in \S\ref{sec:chunk-retrieval-performance}.\looseness-1

In this work, we introduce \method{} (\texttt{L}anguage \texttt{M}odels with annotated \texttt{Ent}ity mentions), a suite of pretrained models, a knowledge-rich pretraining corpus fully annotated with fine-grained entity mentions, and an entity-based retrieval method which allows full transparency into where a specific entity is mentioned in training. The focus on entity mentions stems from mounting evidence that entity names are central to the construction of knowledge representations in LMs \citep[rather than individual tokens;][]{li-etal-2021-implicit, meng2022locating, dissecting, keen}. 

Specifically, the \method{} suite contains: \textbf{(1) A pretraining corpus built on English Wikipedia} with 7.3M fine-grained entity annotations across all documents, derived from three complementary sources: hyperlinks, entity linking, and coreference resolution (\Cref{fig:intro}, left, \S\ref{sec:metadata-gen}). \textbf{(2) An Elasticsearch index with over 10.5M chunks}, enabling retrieval of all pretraining data chunks mentioning a specific entity by its unique Wikidata identifier (\Cref{fig:intro}, center, \S\ref{sec:pretraining}). \method{}'s entity-based retrieval outperforms string-based methods similar to WIMBD \citep{wimbd} and Infinigram \citep{infinigram} on 66.3–80.4\% of entities while maintaining >97\% precision as retrieval depth grows, where string-based precision drops to as low as 27\%. \textbf{(3) A collection of 12 LMs} with 170M, 600M, and 1B parameters, pretrained on the annotated data with 110 intermediate checkpoints per epoch and 4K checkpoints total (\Cref{fig:intro}, right, \S\ref{sec:overview}). Despite only training \method{} models on 0.03\%--4.7\% of the tokens used to train LMs of similar sizes, on factual question answering \cite{popqa}, \method{} models reach comparable performance (7.4\% overall, 66\% on questions about frequently co-occurring entities) to Pythia-1.4B \citep{pythia} (8.7\%, 67\%) and OLMo-1B \citep{olmo1} (10.4\%, 66\%). They do have noticeably lower performance than OLMo-2-1B \citep{olmo2} (13.9\%, 74\%) and SmolLM-1.7B \citep{smollm} (14.9\%, 73\%) mostly due to recall failures on rare facts (\cref{sec:experiments}).

Beyond releasing this resource, we use \method{} to explore several open questions about how pretraining data shapes knowledge in LMs. In \S\ref{sec:plasticity}, we apply \method{} to study knowledge acquisition and forgetting trends, and in-context editing. Because \method{} reveals which training steps introduce chunks mentioning a given entity, we can examine how both fact frequency and location in training correlate with learning and forgetting. We find that LMs learn, and also forget, frequent facts more over the course of training, and that these trends differ across relation types (\cref{subsec:knowledge-acquisition}). Further, leveraging the diverse entity mention annotations, we find that LMs internalize facts better when entities are named explicitly in the data---an analysis infeasible with existing string-based tools (\cref{subsec:entity-mention}). Finally, we show that in-context editing success correlates with the frequency of entity co-occurrence in the data (\cref{subsec:cooccurincontext}) and to the rate of inconsistencies in the model’s predictions over training (\cref{subsec:brittleness}). \looseness-1

Overall, \method{} is a novel tool for precisely tracking information during pretraining, laying the groundwork for studying the interplay between pretraining data and knowledge representations in LMs. We release \method{} to the community as an open-source suite on HuggingFace, including the pretraining data, fine-grained entity mentions, pretrained models, and intermediate checkpoints. For ways \method{} can support research into plasticity, factuality, hallucinations, and pretraining dataset composition, see \cref{sec:discussion}.

\section{Labeling the English Wikipedia with Fine-Grained Entity Mentions}
\label{sec:metadata-gen}
To support analyzing knowledge evolution over training, we select a corpus rich in factual content and annotate its documents with fine-grained entity mentions (\Cref{fig:intro}, left). These annotations enable tracking exactly which entities a model observes at each training step when training on this data. Next, we explain our choice of Wikipedia for pretraining data, and the annotation process.

\paragraph{Pretraining data}
Wikipedia is a natural choice for the source of pretraining data because it is a knowledge base that is structured around entities, such as people, locations, and world events.  
Importantly, there is an integrated system for disambiguating entities using unique identifiers, called \textit{Wikidata QIDs} \citep{wikidata}, and related entity pages are connected through hyperlinks. Moreover, it provides a snapshot of world knowledge at a given time which minimizes contradicting information. Wikipedia is also commonly used to build knowledge benchmarks \citep[e.g.,][]{petroni-etal-2021-kilt, entityquestions, paq, popqa}, making it well-suited for tracking how changes in pretraining affect learning.\looseness-1

\paragraph{Annotation objectives}
Our goal is to be able to query which entities a model saw at any given training step. To achieve this, we built an annotation process that maps character spans in the pretraining data which mention an entity, called \textit{entity mentions}, to unique Wikidata QIDs.\footnote{The annotation pipeline used 8 H100 GPUs (80GB VRAM each). Running coreference resolution and entity linking took 5 days with full GPU utilization.} 
Entity mentions are created for each data chunk, but are not exposed to the model during training—rather, the QIDs are the keys used to retrieve all the data associated with an entity for post-hoc analysis. We design entity mentions to satisfy the following criteria (see \Cref{fig:LMEnt_scoring} for illustration):

\begin{enumerate}[leftmargin=2em, labelindent=0pt, topsep=2pt, itemsep=0pt, label=\textbf{C\arabic*:}, align=left]
    \item Disambiguate between entities with similar names and aliases, e.g., \textcolor{lightgreen}{\textit{``Buffalo''}} for the football team and \textcolor{darkpeach}{\textit{``Buffalo''}} for the city.
    
    \item Capture entities referenced indirectly through descriptive phrases and pronouns, e.g., \textcolor{lightgreen}{\textit{``the team''}}, \textcolor{lightgreen}{\textit{``their''}} and \textcolor{lavender}{\textit{``the hospital''}}.
    
    \item Resolve entity mentions using the full document context
    e.g., \textcolor{darkpink}{\textit{``Allen''}} (Chunk 2) should be annotated with reference to his explicit name, \textcolor{darkpink}{\textit{``Joshua Allen''}} (Chunk 1).
\end{enumerate}

We meet \textbf{C1} and \textbf{C2} by leveraging three complementary sources for annotating entity mentions (\Cref{subsec:entity_mentions}), and \textbf{C3} by applying the scoring process to the entire document before chunking (\Cref{subsec:clustering-entity-mentions}).

\subsection{Entity Mention Sources}
\label{subsec:entity_mentions}

\paragraph{Hyperlinks} identify the first occurrences of explicit entity mentions like Mention (1) in \cref{fig:LMEnt_scoring} which links to the URL of the football team's Wikipedia page, and can be mapped to \textcolor{lightgreen}{\texttt{Q221626}} using Wikidata's Query Service. To extract hyperlinks, we parse the raw XML Wikipedia dump and extract the \texttt{url} attributes from \texttt{href} tags.

\paragraph{Entity linking} is used to increase coverage of direct mentions like \textcolor{darkpeach}{``\textit{Buffalo}''} since hyperlinks offer limited coverage within a document.\footnote{Wikipedia style guidelines recommend linking only the first mention of an entity in a document: \url{https://en.wikipedia.org/wiki/Wikipedia:Manual_of_Style/Linking}.} 
We use ReFinED \citep{refined}, a state-of-the-art modular entity linking system that performs mention detection, fine-grained typing, and entity disambiguation. ReFinED supports zero-shot entity linking by encoding the mention context and candidate entity descriptions using RoBERTa \citep{roberta}, and selecting the entity whose description embedding best aligns with the mention representation. By replacing the underlying base of entity descriptions from our Wikipedia dump, we can identify mentions of new entities and map them to their QIDs without retraining ReFinED.

\paragraph{Coreference resolution} connects indirect mentions, like pronouns, aliases, and generic descriptors, to their entity QIDs.
The two sources above are used to extract explicit entity mentions; however, entities may still be referred to implicitly, e.g., \textcolor{lightgreen}{\textit{``the team''}} in \Cref{fig:LMEnt_scoring}. To fill in this gap, we use the Maverick coreference resolution model \citep{maverick}, which achieves state-of-the-art performance on WikiCoref \citep{wikicoref}. To run Maverick at scale, we reduce its memory footprint by replacing the model backbone, allowing us to parallelize inference on documents (see \Cref{sec:run-maverick} for details).
Maverick outputs clusters of mentions that refer to the same entity. However, these mentions must still be mapped to their QIDs, which we discuss below.

\subsection{Scoring Entity Mentions in a Document}
\label{subsec:clustering-entity-mentions}

Every source yields a set of entity mentions per document in the corpus, and each entity mention is defined by its character span $m = (c^{\text{start}}, c^{\text{end}})$. To consolidate the mentions found across sources, we define source-specific scoring procedures which quantify how confidently a mention can be mapped to a QID (\Cref{fig:LMEnt_scoring}):

\begin{itemize}
[leftmargin=1.5em, labelindent=0pt, topsep=2pt, itemsep=2pt]
\item \textbf{Hyperlinks (H) \pushpin{}} This is the most reliable source since hyperlinks are manually added by human editors. We define $\textbf{H}(m, \textcolor{lightgreen}{\texttt{Q221626}}) = 1$ if there is a hyperlink spanned by $m$ directing to the Wikipedia page associated with \textcolor{lightgreen}{\texttt{Q221626}}, and $\textbf{H}(m, \textcolor{lightgreen}{\texttt{Q221626}}) = 0$ otherwise.\looseness-1
    
\item \textbf{Entity linking (EL) \link{}} The ReFinED model already performs entity disambiguation and provides a score reflecting its confidence that a mention links to a QID, e.g., $M(m, \textcolor{darkpeach}{\texttt{Q40435}}) = 0.98$. Therefore, we simply use this score such that $\textbf{EL}(m, \textcolor{darkpeach}{\texttt{Q40435}}) = M(m, \textcolor{darkpeach}{\texttt{Q40435}})$.

\item \textbf{Coref (C) \leftarrowemoji{}} Suppose we are trying to resolve the QID of the implicit mention, \textcolor{lightgreen}{\textit{``the team''}}. We run Maverick on the document, and it finds that \textcolor{lightgreen}{\textit{``the team''}} is related to the cluster of mentions containing  \textcolor{lightgreen}{\textit{``Buffalo Bills''}}, \textcolor{lightgreen}{\textit{``the Buffalo''}}, \textcolor{lightgreen}{\textit{``the Bills''}}, and \textcolor{lightgreen}{\textit{``their''}}. How can we leverage the cluster to identify the QID of \textcolor{lightgreen}{\textit{``the team''}}? We compute a distribution of scores over all QIDs mapped to some mention in the cluster by the other two sources. In this case, \textcolor{lightgreen}{\texttt{Q221626}} is the only QID supported by this cluster so $\textbf{C}(\textcolor{lightgreen}{\textit{``the team''}}, \textcolor{lightgreen}{\texttt{Q221626}}) = 1.0$. Since this score is computed using the entire cluster, the $\textbf{C}$ score is the same across all mentions in the cluster. Occasionally, a coreference mention encapsulates multiple entities, as in \textcolor{lavender}{``\textit{John R. Oishei Children's Hospital }}\textcolor{darkpeach}{\textit{Buffalo''}}, which leads to a score distribution over several QIDs. In this case of mapping ambiguity, we rely on textual similarity of mentions in the cluster to promote one QID over another. \S\ref{sec:coreference-scoring-appendix} provides the formal definition of this score, and a detailed explanation of how we resolve mapping ambiguity.

\end{itemize}

The final entity mention structure maps each mention $m$ to a list of candidate QIDs with up to three source scores per candidate. This list structure accommodates spans that encompass multiple entities, and multiple scores per candidate allow us to filter chunk retrieval based on source type and confidence thresholds (\S\ref{sec:index-creation}). \S\ref{sec:error-analysis-appendix} includes a qualitative analysis of error patterns over a random sample of 112 mentions, justifying the design of the scoring procedures.

\section{Retrieval from Pretraining Data}
\label{sec:pretraining}

Now that we have entity mentions for every document (\Cref{fig:intro}, left), we need to process the documents into chunks of information for pretraining, and associate the relevant entity mentions with each chunk. By iterating over the entity mentions of each chunk at a particular training step, we can track the entities seen over training (\Cref{fig:intro}, right). \S\ref{sec:chunking} describes how documents are processed into chunks, and \S\ref{sec:index-creation} discusses entity-based retrieval over our index of chunks (\Cref{fig:intro}, center).

\subsection{Data Processing for Training}
\label{sec:chunking}

\paragraph{Tokenization}
Documents are tokenized with the dolma2-tokenizer
\citep{dolma}, resulting in a dataset with 3.6 billion unique tokens. For reference, this corresponds to around 0.09\% of the tokens used to train OLMo-2 \citep{olmo2}.

\paragraph{Chunking}
A chunk is a sequence of tokens that the model independently processes per training step. Unlike existing chunking strategies, like \textit{concat-and-chunk} used in \citet[]{roberta, fairseq, llama1, llama2}, we ensure that each chunk only contains content from a single document using the Variable Sequence Length Curriculum \citep{vsl}. This prevents the model from learning spurious correlations by attending to unrelated documents.

Chunking curricula split documents at some sequence length, which can split entity mentions across chunk boundaries. To prevent this, we modify the chunking logic to terminate just before a mention and pad the remaining tokens to the desired length. The entity mentions for a chunk are defined as those fully contained within it, and we adjust them so they are relative to the chunk's boundaries. 
Practically, we extend the OLMo-Core framework \citep{olmo2} to include entity mentions in the chunk metadata.

\subsection{Entity-Based Chunk Retrieval}
\label{sec:index-creation}
We build an Elasticsearch index \citep{elasticsearch} containing one entry per chunk in the pretraining corpus. Each entry stores the \texttt{chunk\_id}, the chunk text, and its associated entity mentions. As illustrated in \Cref{fig:LMEnt_scoring} and described in \Cref{subsec:clustering-entity-mentions}, each entity mention is linked to a set of candidate entities, where each candidate is identified by its QID and accompanied by three source-specific scores.

To \textbf{retrieve all chunks} that mention a target entity, we query the index for chunks containing at least one mention whose candidate list includes the target entity’s QID (see \Cref{subsec:clustering-entity-mentions}). Retrieval can also be filtered by source type and score thresholds. Retrieval algorithms typically apply an ordering scheme to the items they return \citep{robertson1994okapi}. To accommodate this, we assign a score to each chunk based on the entity mentions that match the retrieval query. For every matching mention, we compute a weighted average of its three source-specific scores. The chunk’s final ranking score is defined as the maximum over the average scores of all its matching mentions.

To \textbf{retrieve all training steps} that mention a target entity, we first retrieve all corresponding chunks from the index, and then map each retrieved chunk to the training step in which it was introduced. This mapping is straightforward since the \texttt{chunk\_id} is the index of the chunk in the dataset, and its associated training step is the index of the batch that contains it in the dataloader.\footnote{There are multiple chunks per batch, determined by a global batch size of 32,648 tokens.}
\section{The \method{} Suite}
\label{sec:overview}

With the procedures for annotating entity mentions and processing pretraining data established, we now provide an overview of the resulting pretraining data index and \method{} models. 

\paragraph{Pretraining data index}

Our pretraining data includes 3.6B tokens over 10.5M chunks, and features more than 7M different entities. The 400M mentions extracted are composed of 115M mentions from hyperlinks, 203M from entity linking, and 310M from coreference resolution. Additional statistics are provided in \S\ref{sec:appendix-pretraining-data-stats}, \Cref{tab:datset_stats}.

\paragraph{Pretrained models}
We introduce a collection of models trained on the dataset, which serve as a testbed for analyzing knowledge evolution over training. Specifically, we train models of three different sizes, 170M, 600M, and 1B parameters, based on the OLMo-2 architecture \citep{olmo2}. 
Notably, the 170M model has the compute-optimal size for 3.6B tokens \citep{chinchilla}. Each model is trained for 1, 2, 4, and 6 epochs (3.6B, 7.2B, 14.4B, 21.6B tokens, respectively), resulting in four variants per size.  Each epoch initializes a different batching and ordering of the chunks. We also release intermediate checkpoints every 1,000 training steps, yielding 110 checkpoints per epoch and proportionally more for longer training schedules. Additional details on model training are in \S\ref{sec:models-appendix}.

\begin{figure}[!t]
\setlength\belowcaptionskip{-9pt}
    \centering
    \includegraphics[width=.85\linewidth]{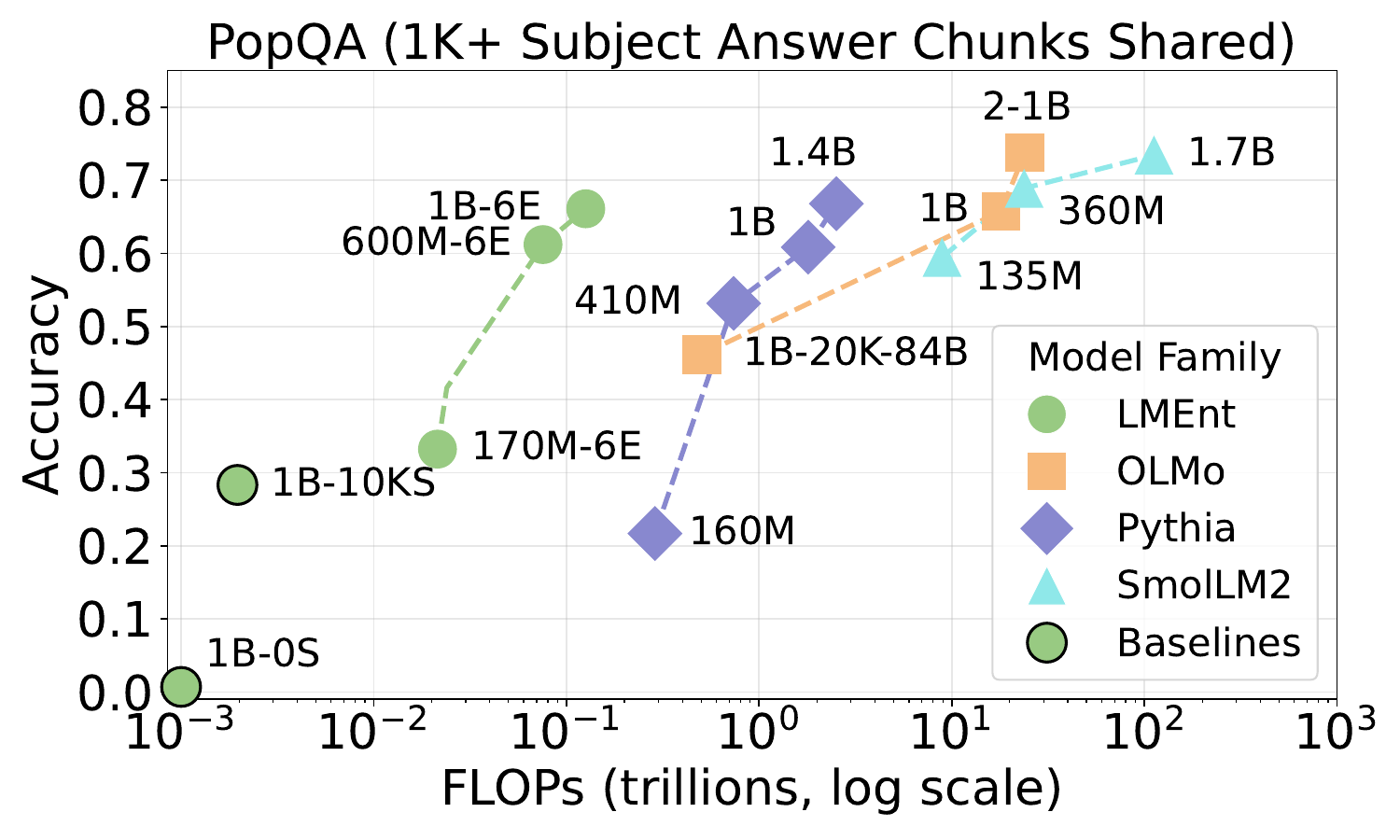}
    \caption{Accuracy on PopQA questions, whose subject and answer entities co-occur frequently, as a function of compute budget. \method{} models achieve comparable performance to other models with better compute efficiency.
    }\looseness-1
    \label{fig:popqa-flops}
\end{figure}

\section{Experiments}
\label{sec:experiments}
In the previous sections, we described the construction and composition of the \method{} suite. Here, we establish the credibility of the suite by showing that \method{} models match the knowledge recall performance of existing open-source models (\S\ref{sec:model-performance}), and that entity-based chunk retrieval using \method{} entity mentions outperforms string-based methods (\S\ref{sec:chunk-retrieval-performance}).

\subsection{Model Performance}
\label{sec:model-performance}

\paragraph{Experimental setup}
To evaluate the \method{} models, we use benchmarks that test for knowledge in Wikipedia. We choose two widely adopted benchmarks: PopQA \citep{popqa} and PAQ \citep{paq}. PopQA contains questions about 11K entities with varying popularities, and PAQ has 65M questions generated from Wikipedia passages. To keep computational costs manageable, we subsample PAQ  and only consider samples containing the same subject entities found in PopQA. This results in an evaluation set of approximately 70K questions.\footnote{Evaluating a 1B parameter model on all PAQ questions using one H-100 80GB GPU would take 200 days.} Results on PAQ are found in \S\ref{sec:popqa-paq-perf-appendix}.
In addition to knowledge-centric tasks, we report results on commonsense reasoning, multiple choice, and reading comprehension tasks in \S\ref{sec:results-not-knowledge}.
Since \method{} models are not instruction-tuned, we convert all samples into cloze-style prompts, such that the answer is the next expected phrase. For example, the original QA-style query ``\textit{What TV network does Funny or Die Presents air on?}'' translates to ``\textit{Funny or Die Presents airs on the TV network}''. Full details of this conversion process are in \S\ref{sec:paq-cloze-appendix}.

\paragraph{Baselines}
We evaluate the performance of \method{} models against leading open-source models of comparable sizes, including Pythia (160M, 410M, 1B, 1.4B) \citep{pythia}, OLMo (1B, 1-20K-84B) \citep{olmo1}, OLMo-2-1B \citep{olmo2}, and SmolLM2 (135M, 360M, 1.7B) \citep{smollm}. We include OLMo-1-20K-84B---an intermediate checkpoint of OLMo-1B trained for 20K steps on 84B unique tokens---because it is the most comparable baseline to our models in terms of training token count.\footnote{Notably, OLMo-1-20K-84B is trained on 4 times more data than our largest \method{}-1B model, and it may not have observed all the facts we evaluate on.} We also include two intermediate \method{} model baselines: \method-1B-0E (randomly initialized) and \method-1B-0.1E (trained for 10K steps). These serve to illustrate how more training data improves knowledge recall capabilities.\looseness-1

\begin{figure}[t]
\setlength\belowcaptionskip{-9pt}
    \centering
    \includegraphics[width=0.65\linewidth]{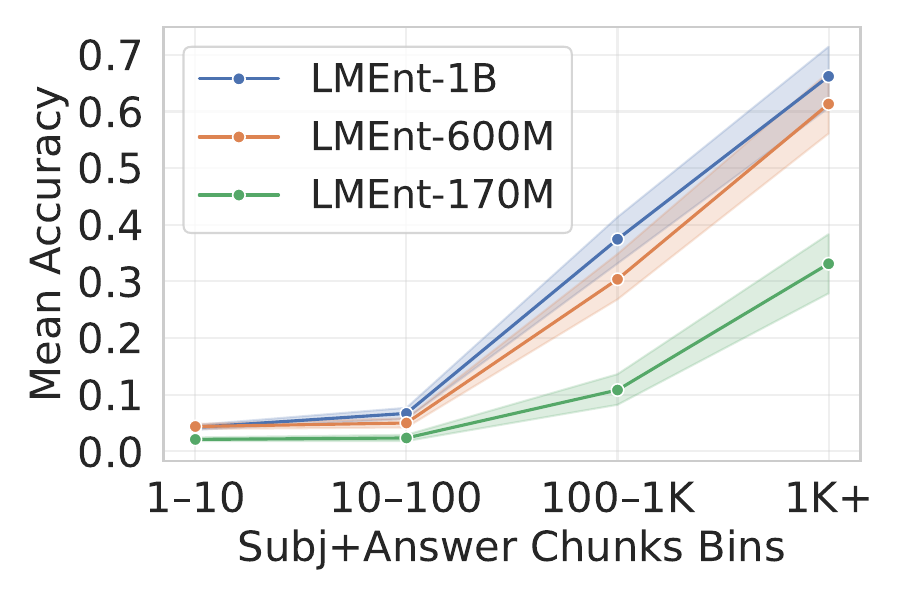}
    \caption{Mean accuracy on PopQA questions binned according to the number of chunks that the subject and answer entities co-occur in. Increasing model size helps learning associations between entities that appear more frequently together.}
    \label{fig:popqa-model-size}
\end{figure}

\paragraph{\method{} models are valid resources for studying knowledge} 

\Cref{fig:popqa-flops} shows the performance of \method{} models and the baselines as a function of compute budget, measured in FLOPs. Despite being trained with two orders-of-magnitude less compute, \method{} models achieve performance comparable to both Pythia and OLMo-1B on questions in PopQA whose subject and answer entities co-occur frequently in the pretraining dataset. We also present results for all frequency levels in \S\ref{sec:popqa-paq-perf-appendix} which show similar trends. However, since PopQA is dominated by questions about tail facts, overall accuracy is lower across all models.

Notably, the \method{}-600M model surpasses the OLMo-1B intermediate checkpoint and the Pythia-1B checkpoint, which reaffirms that pretraining on Wikipedia enables models to acquire substantial factual knowledge \citep{bert} and achieve better compute-performance trade-offs. Though this evaluation aligns with the pretraining objective of \method{} models, while the baseline models are trained on a broader mixture of web, math, and code data, it supports their validity as resources for studying knowledge.

\paragraph{Measuring the effect of model scale on knowledge encoding}

Since all \method{} models are pretrained on the same data, we can isolate the effect of model size on knowledge acquisition \citep{roberts-etal-2025-compute}. \Cref{fig:popqa-model-size} shows that increasing model size improves learning of frequent facts (subject and answer co-occur in $\geq100$ chunks), while having little effect on tail facts (1--100 chunks). We leave exploring how scaling model size further can enhance knowledge learning for future work.

\begin{table}[t]
\footnotesize
\setlength\belowcaptionskip{-9pt}
\setlength\tabcolsep{3.8pt}
    \centering
    \resizebox{\linewidth}{!}{
    \begin{tabular}{lcp{6.5cm}}
    \toprule
         & Retrieval & Match Examples  \\
         \midrule
         \multirow{5.0}{1em}{\rotatebox{90}{\method}} & QID + Scores & {\texttt{Buffalo, New York; Buffalo; the Queen City; this city; the City's; the City of Buffalo; the city Buffalo; a town on the banks of Lake Erie; his hometown}} \\
         \midrule 
         \multirow{5.0}{1em}{\rotatebox{90}{CS-SS}} & {Canonical} & {\texttt{Buffalo, New York}} \\ \cmidrule{2-3}
          & \multirow{2}{*}{Extended} & {\texttt{Buffalo, New York; Buffalo, NY; Buffalo, N. Y.; City of Light; The Queen City; The Nickel City}}\\
         \midrule 
         \multirow{5.0}{1em}{\rotatebox{90}{CI-SS}} & {Canonical} & {\texttt{buffalo, new york}}{\texttt{}}\\ \cmidrule{2-3}
          & \multirow{2}{*}{Extended} & {\texttt{buffalo, new york; buffalo, ny; buffalo, n. y.; city of light; the queen city; the nickel city}}\\      
          \bottomrule
    \end{tabular}
    }
    \caption{Matched entity mentions of \textcolor{darkpeach}{``\textit{Buffalo, New York}''} for different retrieval methods. Entity-based retrieval over the \method{} index matches on \textcolor{darkpeach}{\texttt{Q40435}}. The \texttt{Canonical} method matches only the entity’s canonical name, whereas the \texttt{Extended} method matches any of its aliases.}
    \label{tab:mention-matches}
\end{table}

\subsection{Chunk Retrieval Performance}
\label{sec:chunk-retrieval-performance}
A key use-case of the \method{} index is retrieving all the pretraining data chunks that mention a specific entity. Here, we compare the quality of entity-based retrieval (\Cref{sec:index-creation}) to existing string-based retrieval methods. 

\paragraph{Experimental setup}
We evaluate on a test set of 1K entities, chosen via stratified sampling by their number of hyperlinks.
We use the entity-based retrieval over the \method{} index to retrieve their corresponding chunks, matching both on entity QID and on at least one of the score thresholds: $\textbf{H} = 1$, $\textbf{EL} \geq 0.6$, $\textbf{C} \geq 0.6$. These thresholds were empirically determined using a development set of 60 entities, described in \S\ref{sec:score-thresholds-appendix}. We then compare the set of retrieved chunks to those retrieved by popular string-based methods. 

To evaluate retrieval precision, we use Gemini 2.5 Flash \citep{gemini} as a judge \citep[]{llm-judge-survey, llm-judge-1} which predicts ``Yes'' or ``No'' based on whether the mention directly relates to the target entity. The prompt used and the statistical test justifying our use of an LLM-judge are provided in \S\ref{sec:gemini-prompt-appendix}. Since some entities are mentioned very frequently (\Cref{fig:retrieval-histogram}), we randomly sample 100 chunks from the set retrieved per method and entity, and measure the absolute number of chunks judged as ``Yes'' for each entity.

\begin{figure}[!t]
\setlength\belowcaptionskip{-9pt}
    \centering
    \includegraphics[width=\linewidth]{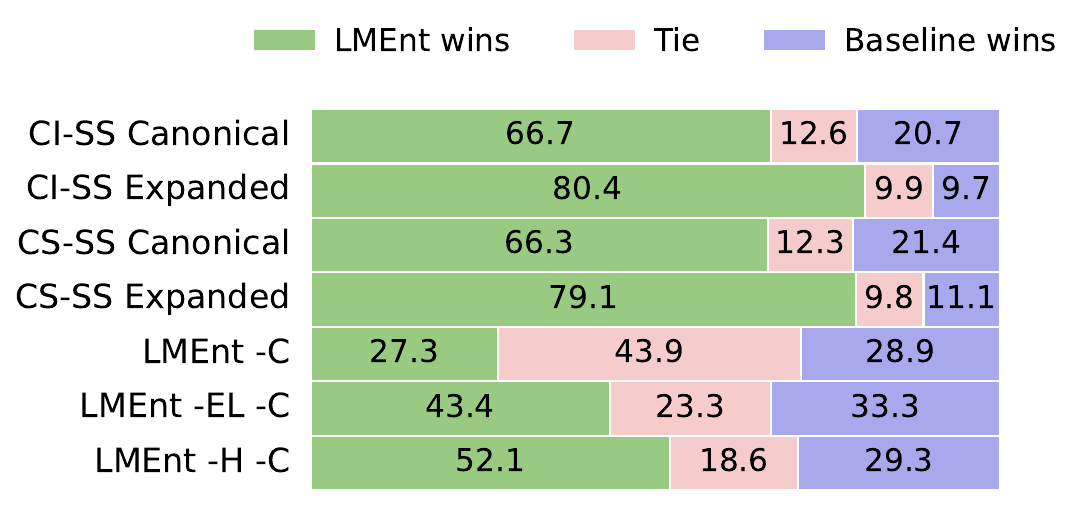}
    \caption{Pairwise wins rate for entity-based retrieval (\method{}) with multiple string-based methods and ablations. \method{}'s entity-based retrieval outperforms string-based methods by 66.7\%--80.4\%. Ablations (bottom three rows) show that hyperlinks and entity linking are the most crucial sources of entity mentions.}
    \label{fig:win-rate-summary}
\end{figure}

\paragraph{Baselines}
We compare to string-based baselines that retrieve chunks using either case-sensitive (\texttt{CS-SS}) or case-insensitive (\texttt{CI-SS}) exact matches of entity names within the chunk text. The \texttt{Canonical} variant matches only on the entity's canonical name, and the \texttt{Extended} variant additionally matches against the entity’s Wikidata aliases which is a common strategy used to improve recall \citep[e.g.,][]{ripple, multihop}.

The \texttt{CS-SS Canonical} baseline approximates the Infinigram \citep{infinigram} tool which matches based on case-sensitive n-grams, and \texttt{CI-SS Canonical} baseline approximates WIMBD \citep{wimbd} which retrieves on exact case-insensitive string matches. Although \texttt{CS-SS Canonical} relies on exact string matches rather than n-gram matches, it is a stronger baseline since shorter references introduce noise due to alias ambiguity (e.g., the n-gram ``\textit{Buffalo}'' indiscriminately matches chunks that refer to the city, animal, or football team). \Cref{tab:mention-matches} summarizes the retrieval strategies used by \method{} and the baselines. For clarity, we also include examples of mentions that triggered the retrieval of chunks for \textcolor{darkpeach}{\texttt{Q40435}}.\looseness-1

\begin{figure}[!t]
\setlength\abovecaptionskip{5pt}
    \centering
    \includegraphics[width=0.8\linewidth]{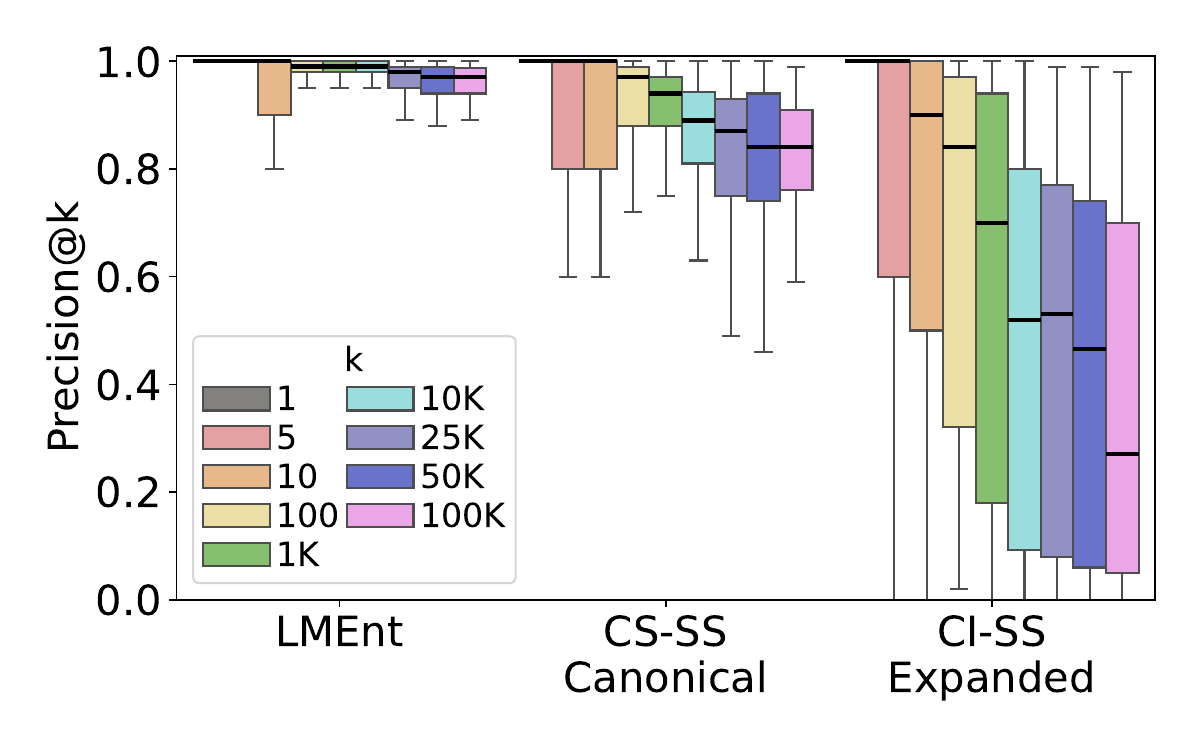}
    \caption{Precision at various retrieval depths ($k$), the top-$k$ chunks retrieved per entity. As depth increases, \method{} maintains $>97\%$ precision, while \texttt{CS-SS Canonical} and \texttt{CI-SS Expanded} decrease to 84\% and 27\%.} 
    \label{fig:precision-at-k}
\end{figure}

\paragraph{Entity-based retrieval largely outperforms the baselines}
\label{sec:index-eval}

\Cref{fig:win-rate-summary} presents pairwise win rates between retrieval methods, where a win is defined as one method retrieving more ``Yes''-judged chunks than another for a given entity. Entity-based retrieval (\method wins) consistently outperforms all other methods, retrieving more correct mentions for 66.3\%--80.4\% of the entities. Surprisingly, the \texttt{Expanded} variants of both \texttt{CI-SS} and \texttt{CS-SS} perform worse than their canonical counterparts by $40\%$, suggesting that additional noise introduced by alias expansion outweighs any gains in recall (\S\ref{sec:ss-win-rate-appendix}, \Cref{fig:win-rate-summary-appendix}). Additionally, we observe that entity-based retrieval is superior to string-based methods across all entity frequency levels---notably on tail and torso entities which together make up 99.7\% of the entities in Wikipedia---and often by a substantial number of chunks (\S\ref{sec:retrieval-rare}, \Cref{fig:retrieval-histogram}; \S\ref{sec:win-rate-detailed-appendix}, \Cref{fig:win-rate-distribution}).

\paragraph{Ablating entity mention sources reduces retrieval quality}
We also ablate sources of entity mentions in entity-based retrieval and examine the impact on win rate (\Cref{fig:win-rate-summary} last three rows). Hyperlinks are the most valuable component as relying solely on them (\texttt{\method{} -EL -C}) wins on 33.3\% of entities. In contrast, using only entity linking (\texttt{\method{} -H -C}) lowers wins to 29.3\%. This indicates that while hyperlinks play a crucial role, entity linking alone remains reasonably effective---highlighting the potential of applying \method{}'s annotation pipeline to pretraining corpora beyond Wikipedia, where hyperlinks are unavailable.
Removing coreference resolution (\texttt{\method{} -C}) has little effect, because judging chunks with only implicit mentions is difficult. For example, when Donald Trump is referred to solely as ``\textit{his Mexico City Policy}'', it is difficult to identify that ``\textit{his}'' implicitly refers to Trump without the context earlier in the document which mentions him explicitly.

\begin{figure}[!t]
\setlength\abovecaptionskip{5pt}
    \centering
    \includegraphics[width=.8\linewidth]{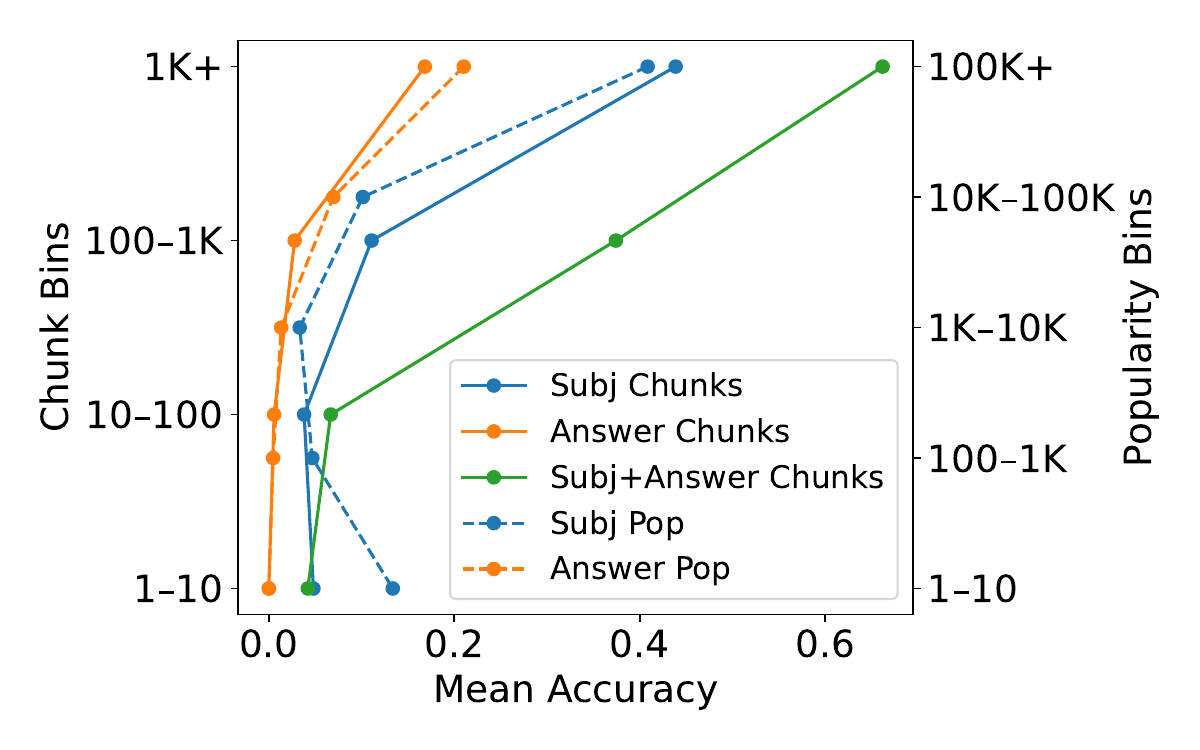}
    \caption{Accuracy for \method-1B on PopQA. Subj+Answer Chunks counts chunks that mention both the subject and answer entities of a question. Answer Chunks and Subject Chunks count chunks that mention the answer and subject entities individually. Subj Pop and Answer Pop are pageview popularities from \citet{popqa}. Subj+Answer Chunks correlates best with the model's performance.}
    \label{fig:m_faithful}
\end{figure}

\paragraph{\method{} maintains high precision as the number of retrieved chunks increases}
We also analyze the precision of retrieved chunks at varying retrieval depths, $k$. For each method, we consider only the top-$k$ chunks retrieved per entity, and randomly sample 100 chunks if $k \geq 100$, for feasibility. An LLM then judges the mention, and precision is computed as the fraction of "Yes" responses out of $k$. As shown in \Cref{fig:precision-at-k}, \method{}'s entity-based retrieval maintains high precision $\geq 97\%$ as $k$ increases, while precision substantially declines for all other methods, reaching $27\%$ and $84\%$ for \texttt{CI-SS} and \texttt{CS-SS}, respectively. This indicates that as more chunks are retrieved, string-based methods introduce noise while \method{} maintains high quality.\looseness-1

\begin{figure*}[!t]
\setlength\belowcaptionskip{-7pt}
\centering

\begin{subfigure}[t]{0.33\linewidth}
  \centering
  \raisebox{14pt}{%
    \includegraphics[width=\linewidth]{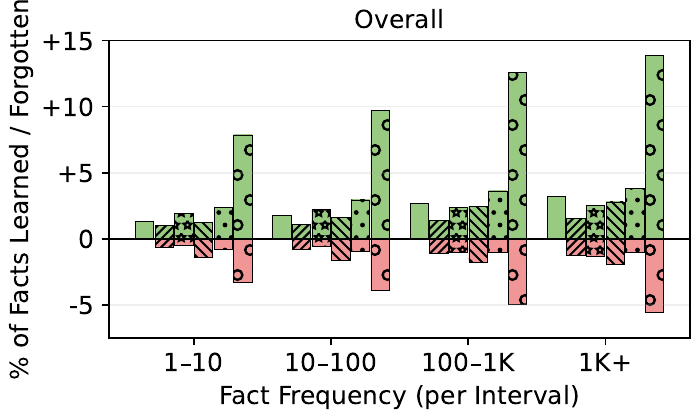}
  }
\end{subfigure}
\hfill
\begin{subfigure}[t]{0.31\linewidth}
  \centering
  \includegraphics[width=\linewidth]{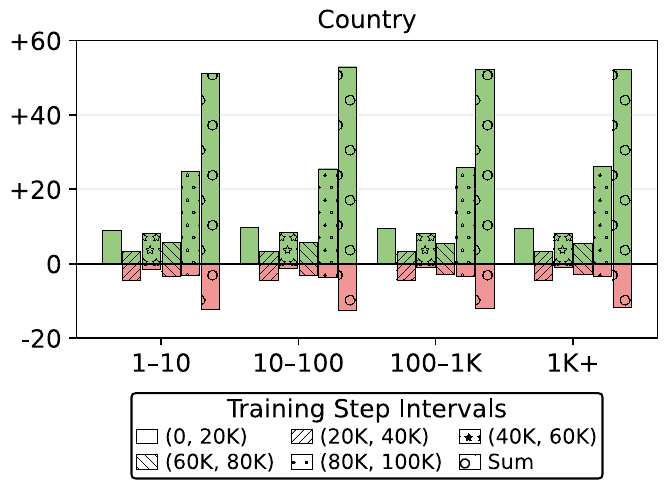}
\end{subfigure}
\hfill
\begin{subfigure}[t]{0.31\linewidth}
  \centering
  \raisebox{22pt}{%
    \includegraphics[width=\linewidth]{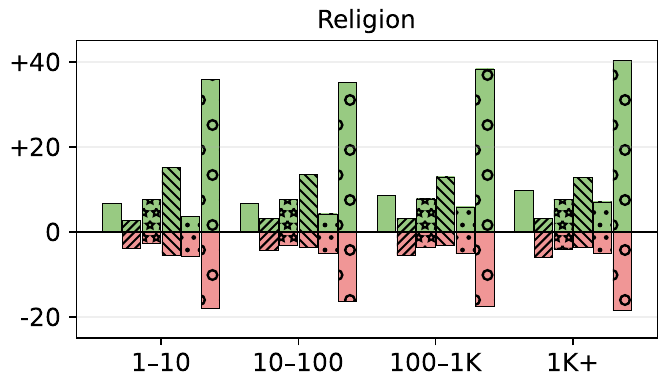}
  }
\end{subfigure}

\caption{Percentage of facts \textcolor{learned}{learned} and \textcolor{forgotten}{forgotten} by fact frequency between intermediate checkpoints of \method-1B.
Overall (left) shows that while fact frequency correlates with knowledge gains, both learning and forgetting increase with frequency. The learning and forgetting trends for specific Country (middle) and Religion (right) relations differ across training step intervals. Overall learning is lower since the model struggles to answer questions about certain relations (e.g., Screenwriter in \cref{fig:delta}). }
\label{fig:pos-mg}
\end{figure*}

\paragraph{Entity co-occurrence better aligns with model performance than popularity}
The number of Wikipedia pageviews for a subject entity, denoted as Subj Pop, has been the standard proxy for real-world popularity and is known to correlate with a model’s ability to recall facts about it \citep{popqa}. 
In \Cref{fig:m_faithful}, we compare model performance against several potential indicators. In PopQA, each question concerns a (Subject, Answer) entity pair. Leveraging \method{}'s entity-based retrieval, which can retrieve chunks containing many co-occurring entities, we introduce a new indicator: Subject+Answer Chunks. This is defined as the number of chunks in which the subject and answer entities co-occur. This co-occurrence shows the strongest correlation with model performance. Interestingly, this trend is also consistent for models trained on pretraining data that extend beyond Wikipedia (\S\ref{sec:other-model-indicators}, \cref{fig:other-model-indicators-all}).\looseness-1

\section{Applications of \method{}}
\label{sec:plasticity}

We demonstrate the utility of \method{} by focusing on two fundamental applications: knowledge acquisition and in-context knowledge editing. Specifically, we study learning and forgetting trends over training (\cref{subsec:knowledge-acquisition}), how entity mention forms relate to learning (\cref{subsec:entity-mention}), and how in-context knowledge editing relates to entity co-occurrence frequency (\cref{subsec:cooccurincontext}) and inconsistencies in model predictions over training (\cref{subsec:brittleness}).

\subsection{Knowledge Acquisition in Pretraining}
\label{subsec:knowledge-acquisition}

We study at which point during training models acquire factual knowledge and how this relates to fact frequency, which has been previously approximated using document frequency \citep{pmlr-v202-kandpal23a}, term co-occurrence \citep[]{elazar2023measuringcausaleffectsdata, merullo2025on} and n-gram \citep{wang2025generalization} frequencies. In contrast to prior work, we use \method{} entity mentions, which provide more accurate estimates of fact frequency compared to string-based methods.

\paragraph{Experiment} We evaluate the \method{}-1B model every 20K steps on PopQA. We define a fact by a (Subject, Answer) entity pair, which has been shown to be an effective fact proxy \citep{elsahar-etal-2018-rex}. A fact is considered ``learned'' if the model correctly answers at least one question linking the subject to the answer. Between training steps ($i, i{+}20\text{K}$), we consider all the facts seen in the training step interval and measure (a) \textit{fact frequency}: the number of chunks in this interval where the subject and answer co-occurred, (b) \textit{\% of learned facts}: facts that were not learned at step $i$  and then learned at step $i{+}20\text{K}$, and (c) \textit{\% of forgotten facts}: facts that were learned at step $i$ and then not learned, i.e. ``forgotten'', at step $i{+}20\text{K}$.

\paragraph{Findings} \cref{fig:pos-mg} (left) shows the percentage of facts learned and forgotten as a function of both fact frequency and the training step interval in which the fact was introduced. Focusing on trends across fact frequencies (left, Sum), the model recalls frequent facts from memory more effectively, which is consistent with prior works \citep[]{elazar2023measuringcausaleffectsdata, yu-etal-2023-characterizing, liu-etal-2025-tracing}. However, we also see that LMs forget very frequent facts (1K+, -5.2\%), and successfully learn rare facts (1-10, +8\%). This implies that fact frequency alone does not dictate learning and forgetting dynamics, and may be confounded by other factors like when a fact is introduced in training (discussed next), or its entity mention forms (see \cref{subsec:entity-mention}).

\begin{figure}[!t]
    \centering
    \includegraphics[width=0.75\linewidth]{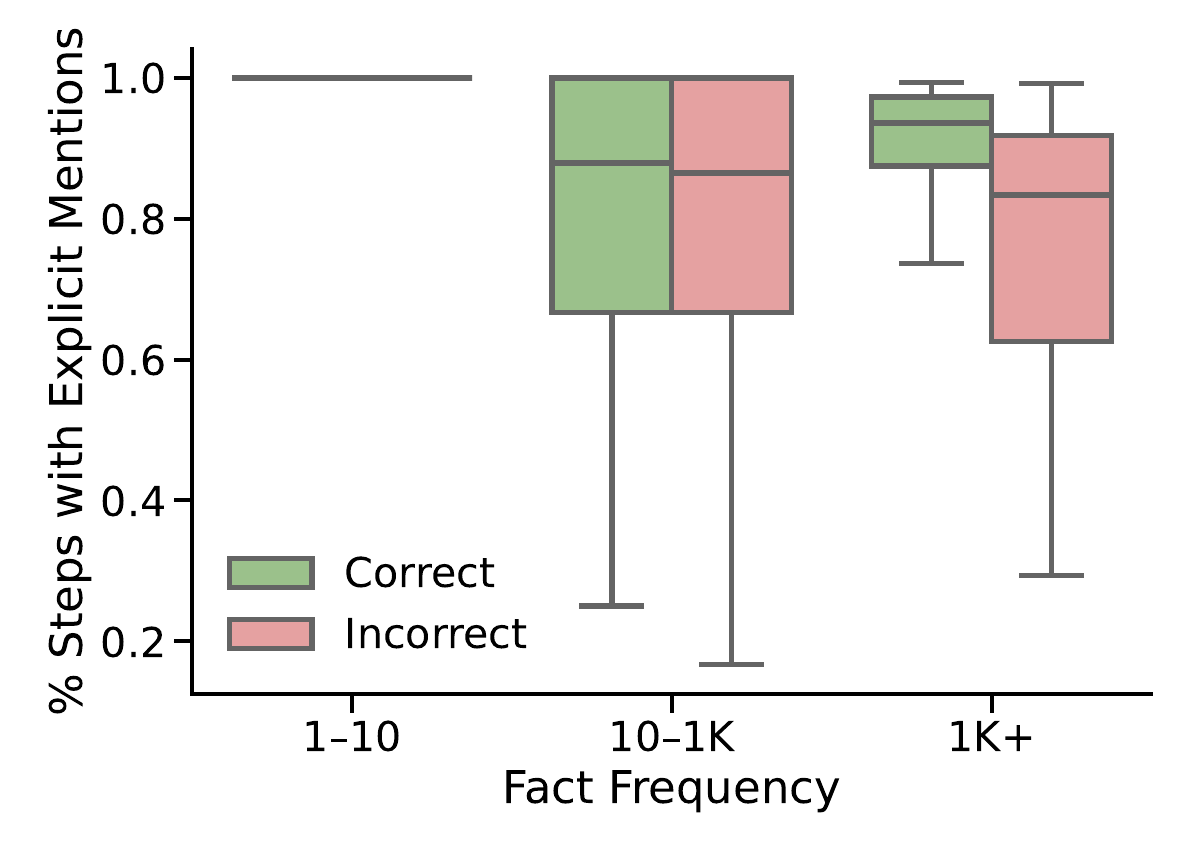}
    \caption{\method-1B is more accurate when entities co-occur in their explicit forms, implying that entity mention forms influence learning.}
    \label{fig:frequent-fact-analysis}
\end{figure}

To analyze how the timing of a fact's introduction affects its acquisition, consider any fact frequency bucket in \cref{fig:pos-mg} (left). We see that the model acquires knowledge in the first 20K training steps, and continues to learn and forget facts as training proceeds, with the highest percentage of facts learned in the last 20K training steps. Moreover, knowledge acquisition is not correlated with the cosine learning rate schedule, which decreases after 1K warmup steps, as acquisition is highest in the last training step interval where the learning rate is lowest.

Further, we see varied learning dynamics when restricting this analysis to specific relations: Country (middle) and Religion (right). For example, the training step interval with highest learning is (80K, 100K) for country, but (60K, 80K) for religion. This observation aligns with previous findings that knowledge recall trends over training differ across relations and entity types \citep[]{multihop,balesni2025lessonsstudyingtwohoplatent}. Results for additional relations can be found in \S\ref{appendix:delta}.

\subsection{Entity Mention Forms and Learning} 
\label{subsec:entity-mention}
The previous experiment shows that fact frequency alone does not fully explain learning and forgetting trends. So far we treated all co-occurrences of the subject and answer entities as equally informative, but entities can be mentioned with various entity mention forms, e.g., ``\textit{Donald Trump}'' vs. ``\textit{the 47th president}''. Here, we leverage the fine-grained \method{} entity mentions to examine the relation between how an entity is mentioned in the pretraining data and the model's ability to learn facts about it. Notably, such analysis is infeasible with string-based methods since they suffer from alias ambiguity. 

\paragraph{Experiment}
We restrict our analysis to chunks in which a (Subject, Answer) entity pair co-occur, and determine whether both the subject and answer entities are mentioned explicitly (e.g., canonical names like ``Donald Trump'') or implicitly through aliases (e.g., ``\textit{the 47th president}'') or pronouns (e.g., ``\textit{he}''). For each (Subject, Answer) entity pair, we then compute the proportion of training steps that introduce chunks which explicitly mention both entities.

\paragraph{Findings} \cref{fig:frequent-fact-analysis} shows the percentage of training steps that mention both the subject and answer entities explicitly, grouped by fact frequency and correctness. For rare facts (1–10), which appear in more focused contexts, all of the training steps contain explicit mentions regardless of correctness, indicating that a different factor is at play (e.g., location in training). For more frequent facts (10-1K), correctly answered questions are associated with a slightly higher proportion of steps with explicit mentions, and this gap is more pronounced for highly frequent facts (1K+). This suggests that mention forms affect how models learn factual knowledge, and may contribute to why LMs forget frequent facts. We encourage future work using \method{} to systematically study these effects, including causal interventions that replace implicit mentions with explicit ones to isolate impacts on learning and forgetting.\looseness-1

\subsection{Entity Co-occurrence and In-Context Editing Success}
\label{subsec:cooccurincontext}

We saw that the co-occurrence of subject and answer entities correlates with the model's performance on related queries (\cref{subsec:knowledge-acquisition}). Here, we extend this analysis to study how co-occurrence statistics affect the model’s ability to edit its learned knowledge in-context \citep[]{zheng2023can, yu-etal-2023-characterizing, ripple}.

\paragraph{Experiment}
We evaluate checkpoints of the \method{}-1B model trained for six epochs that were saved every 100K training steps. Focusing on the questions in PopQA that were answered correctly by at least one of these checkpoints, we generate up to 50 counterfactual answers per question by retrieving all the chunks that mention the subject entity and choosing counterfactual entities that (i) match the original answer type and (ii) co-occur with the subject in the pretraining data. For example, for the fact ``\textit{The capital of France is Paris}'', the city ``\textit{Rome}'', which also co-occurs with ``\textit{France}'' in training, can serve as a counterfactual answer. To evaluate in-context knowledge editing, we prompt every checkpoint with the edited fact followed by the original cloze-style prompt, e.g., ``\textit{The capital of France is Rome. The capital of France is}'', and consider editing successful if the LM outputs the counterfactual answer e.g., ``\textit{Rome}''.

\begin{figure}[!t]
    \centering
    \includegraphics[width=0.8\linewidth]{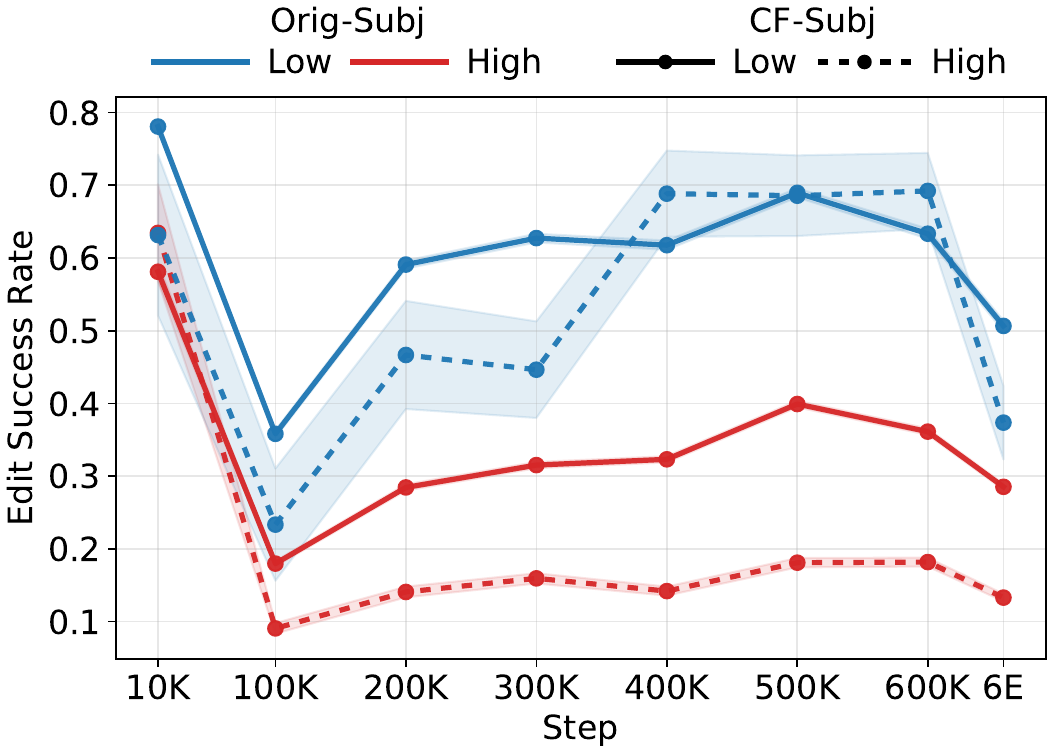}
    \caption{In-context editing success across training checkpoints, split by the co-occurrence frequencies of \texttt{Orig–Subj} and \texttt{CF–Subj} pairs. LMs are more rigid to editing facts with strong associations with either pair (red $<$ blue, dash $<$ solid). \texttt{6E} refers to the final model at epoch six.}
    \label{fig:cf_edit_success}
\end{figure}

\paragraph{Findings} 
~\cref{fig:cf_edit_success} shows in-context editing success over checkpoints, split by the co-occurrence frequencies of two entity pairs: \texttt{Orig–Subj}, the (Subject, Original Answer) entity pair like (``\textit{France}'', ``\textit{Paris}''), and \texttt{CF–Subj}, the (Subject, CF Answer) entity pair like (``\textit{France}'', ``\textit{Rome}''). Each pair is labeled low-frequency (1–10 training steps) or high-frequency (100+), and an edit is categorized by both its \texttt{Orig–Subj} and \texttt{CF–Subj} frequency labels, e.g., the dashed blue line shows edits with low \texttt{Orig–Subj} and high \texttt{CF–Subj}.

We find that \texttt{Orig–Subj} co-occurrence plays the dominant role—when the original answer and subject are weakly associated (blue), in-context editing success is highest. \texttt{CF–Subj} co-occurrence also matters—when the counterfactual entity and subject are weakly associated (solid), the LM more readily adopts the new relation.
Examining trends over training, we see high edit success (58\%-78\%) for the first checkpoint (10K training steps). After approximately one epoch, the model's knowledge becomes rigid for its strong associations (red/dashed). However, for facts with weaker associations, the model exhibits a non-monotonic pattern: editability decreases after the first epoch, partially recovers mid-training, and then declines again at the end. This behavior aligns with prior work showing that continual learning is more effective from intermediate checkpoints \citep{springer2025overtrained}.

\begin{figure}[!t]
    \centering
    \includegraphics[width=0.75\linewidth]{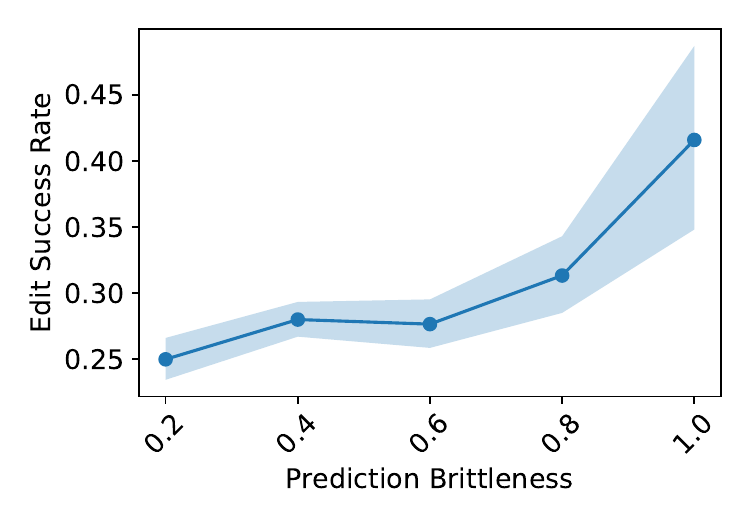}
    \caption{In-context editing success of the \method{}-1B model increases with prediction brittleness. Logistic regression found a monotonic relationship $(\beta = 0.54, z = 4.97, p =6.65e^{-07})$.}
    \label{fig:brittlenesss}
\end{figure}

\subsection{Model Brittleness and In-Context Editing}
\label{subsec:brittleness}
Here, we examine how  inconsistencies in the model's predictions across training relate to the model’s ability to edit its knowledge in-context.

\paragraph{Experiment} 
We evaluate the final checkpoint using the same dataset as in \cref{subsec:cooccurincontext}. We define \textit{prediction brittleness} as the frequency of ``correctness flips'' (Correct$\rightarrow$Incorrect and Incorrect$\rightarrow$Correct) across six checkpoints, sampled every 100K training step, with steps beginning from the first third of training to exclude noisy early-stage prediction fluctuations. Examples whose correctness never changes are omitted as they do not exhibit prediction brittleness.

\paragraph{Findings}
\cref{fig:brittlenesss} shows that in-context editing success increases with prediction brittleness: facts whose answers fluctuate more during training are easier to edit. A binomial logistic regression analysis confirms the relation is monotonic and positive. This result highlights a connection between training dynamics and the editability of the final model, motivating future work on what drives such brittleness and the relationship between knowledge stability and model safety.

\section{Related Work}

\paragraph{Analysis of data acquisition in LMs}
Prior work on how pretraining data shapes factual knowledge has largely relied on synthetic setups, due to limited metadata available for pretraining corpora. These include injecting synthetic facts \citep{chang-etal-2024-how}, constructing artificial biographies \citep{physicsofllm, zucchet2025how}, pretraining on structured fact triplets \citep{kassner-etal-2020-pretrained, dmtriplets}, and identifying knowledge circuits using synthetic data \citep{ou-etal-2025-llms}. In contrast, \method{} enables controlled investigation in realistic settings using natural data.

Other lines of work examine the influence of individual training examples on learning \citep{han-etal-2020-explaining, Hammoudeh2022TrainingDI}, localize where knowledge is encoded in model parameters \citep{kim2025knowledge}, and analyze which properties of pretraining data give rise to hallucinations \citep{sun2025why, zucchet2025how}. These approaches can benefit from \method{}, which allows direct comparison of model checkpoints before and after seeing specific facts.

\paragraph{Retrieval over pretraining corpora}
\method{}'s entity-based retrieval outperformed string-based retrieval methods (\S\ref{sec:chunk-retrieval-performance}) which rely on exact matches \citep{wimbd, olmotrace} or n-gram matches \citep{infinigram}. Many works that study the connection between pretraining data statistics and knowledge acquisition \citep{razeghi-etal-2022-impact, pmlr-v202-kandpal23a, elazar2023measuringcausaleffectsdata, kang-choi-2023-impact, merullo2025on, wang2025generalization} can use \method{}'s entity-based retrieval to refine their findings.

\paragraph{Open-source LM toolkits} Similar to \method{}, Pythia \citep{pythia}, OLMo \citep{olmo1, olmo2}, LLM360 \citep{llm360}, SmolLM \citep[]{smollm, smollm3}, and NVIDIA \cite[]{nemo, nvidia2025nvidianemotronnano2} released a suite of models along with their training data and training framework to facilitate the study of scaling laws and training dynamics in relation to the pretraining data. Other works train their own collection of models to study scaling laws for data curation \citep{magnusson2025datadecide} and specific downstream tasks \citep{bhagia2025establishing}. 
\method{} differs from prior work by providing a practical toolkit for studying factual knowledge in LMs, with a particular emphasis on easily attributing when and where specific information appears in training.

\section{Conclusion and Discussion}

We introduce \method{}, a suite of 12 LMs, pretraining data fully annotated with entity mentions, and an entity-based retrieval index that facilitates studying the connection between knowledge acquisition and pretraining data. We show that \method{} models are capable of knowledge recall tasks, and that our entity-based retrieval outperforms string-based methods on 66.3\%--80.4\% of entities.
Overall, \method{} serves as a flexible and extensible testbed for investigating a broad set of questions regarding knowledge and representations in LMs. In the following sections, we list future applications of \method{} and limitations.

\subsection{Future Applications of \method{}}
\label{sec:discussion}

\paragraph{Knowledge plasticity}

\method{} can be used to study the plasticity of knowledge in language models: identifying steps during pretraining when models are more receptive to acquiring new knowledge. This follows prior findings that pretrained models struggle to learn new facts \cite{lyle2022understanding}, excessive pretraining can make models resistant to fine-tuning \cite{springer2025overtrained}, and LMs store a finite amount of parametric knowledge \citep{allen-zhu2025physics}.

\paragraph{Improving factuality of LMs} Since \method{} provides entity mentions for each chunk in the pretraining corpus, a possible extension is to leverage them to improve the model's factuality and mitigate hallucinations. For example, one could experiment with different data ordering methods, like grouping together all the chunks mentioning the same entity, or augmenting the pretraining data by replacing implicit mentions with explicit entity names.\looseness-1

\paragraph{Effect of other data sources} 
In this work, we train models on a relatively small and knowledge-rich corpus, which results in LMs that are capable in knowledge tasks, yet perform poorly on out-of-distribution tasks, such as commonsense reasoning (\S\ref{sec:results-not-knowledge}). Modern LMs are trained on much larger corpora, often surpassing 10T tokens \cite{smollm}, derived from sources that are \emph{knowledge-poor}, e.g., coding \cite{rozière2024codellamaopenfoundation, hui2024qwen25codertechnicalreport}, synthetic stories \cite{eldan2023tinystoriessmalllanguagemodels}, and formal languages \cite{hu-etal-2025-circuits}. 
The \method{} pretraining dataset can be extended with such sources to analyze if the addition of these tokens improves factuality. The annotation pipeline can also be applied to other knowledge-rich sources that lack hyperlinks, since relying just on entity linking is still effective (\texttt{\method{} -H -C}, \cref{fig:win-rate-summary}).\looseness-1

\paragraph{Mechanistic interpretability} The enhanced visibility that \method{} provides into the training process facilitates a controlled, yet natural, setup for studying the formation of latent knowledge representations and circuits in LMs.

\subsection{Limitations}

\paragraph{Pretraining data, model size and architecture}
\method{} presents a relatively small pretraining corpus and collection of compute-efficient models. Such experimental settings have been successful in recent years, enabling academia to develop procedures that were later scaled effectively, e.g., \cite{rafailov2023direct}. As mentioned in \S\ref{sec:discussion}, an interesting future direction is to scale \method{} annotations to other sources---thereby increasing the size of the pretraining corpus and facilitating studies into how different types of data affect knowledge acquisition. Also, \method{} can be easily extended to architectures beyond dense transformers like mixture-of-experts \cite{dai2024deepseekmoeultimateexpertspecialization, muennighoff2025olmoeopenmixtureofexpertslanguage, Cai_2025}, which is already supported by the OLMo framework \cite{olmo1, muennighoff2025olmoeopenmixtureofexpertslanguage}.

\paragraph{Beyond pretraining}
Since models see the vast majority of tokens during pretraining, this stage is critical for understanding how they acquire and represent knowledge. While our study focuses on pretraining, it is typically followed by mid- and post-training phases. In these stages, models are presented with additional high-quality next-token prediction data and fine-tuning examples \cite{olmo2, kumar2025llmposttrainingdeepdive}. Future work could extend \method{} by adding annotations to the data used for mid- and post-training, and further train \method{} models on this data.

\paragraph{Dense retrievers}
In this work, we do not compare \method{} retrieval against dense retrievers such as DPR \citep{dpr} and ColBERT \citep{colbert}, since sparse retrieval methods like WIMBD \citep{wimbd} and Infinigram \citep{infinigram} are the predominant approaches for retrieving relevant information in pretraining corpora.

\section*{Acknowledgments}
We are grateful to Maor Ivgi, Alon Mendelson, Yanai Elazar, and Ohav Barbi for their valuable discussions and feedback.
We also thank Noam Steinmetz, Clara Suslik, and Asaf Avrahamy for their participation in the LM-judge evaluation. This work was supported in part by AMD's AI \& HPC Fund, the Mila P2v5 grant, the Mila-Samsung grant, the Google PhD Fellowship program, the Alon scholarship, and the Israel Science Foundation grant 1083/24.

\bibliography{tacl2021}
\bibliographystyle{acl_natbib}

\iftaclpubformat
\fi

\onecolumn

\appendix

\section{Appendix: Supplementary Results}

\subsection{Pretraining Data Statistics}
\label{sec:appendix-pretraining-data-stats}
\Cref{tab:datset_stats} summarizes statistics over the \method{} pretraining corpus. These results support the overview provided in \S\ref{sec:overview}.

\begin{table*}[ht]
\setlength{\textfloatsep}{-2pt}
\setlength{\abovecaptionskip}{2pt}
\setlength{\belowcaptionskip}{-8pt}
\setlength{\tabcolsep}{2pt}
\renewcommand{\arraystretch}{0.9}
\scriptsize
\centering
\resizebox{\textwidth}{!}{%
\begin{tabular}{ccc ccc ccccc ccc}
\toprule
\# tokens & \# chunks & \# entities & 
\# batches & \# mentions & \# hyperlinks & 
\multicolumn{5}{c}{\# entities by $H$} &
\# entity linking & \# coref. mentions & \# coref. clusters \\
\cmidrule(lr){7-11}
 & & & (steps) & total & ($H$) &
$1$ & $(1,\,10]$ & $(10,\,100]$ & $(100,\,1\text{K}]$ & $>1\text{K}$ & & & \\
\midrule
3.6B & 10.5M & 7.3M &
109K & 400M & 115M &
993K & 2.2M & 967K & 141K & 13K & 203M & 151M & 310M \\
\bottomrule
\end{tabular}
} 
\caption{Statistics of the \method{} pretraining corpus.}
\label{tab:datset_stats}
\end{table*}

\subsection{PopQA and PAQ Performance}
\label{sec:popqa-paq-perf-appendix}
\begin{figure}[ht]
\setlength{\abovecaptionskip}{4pt}
\setlength{\belowcaptionskip}{-20pt}
    \centering
    \includegraphics[width=.3\linewidth]{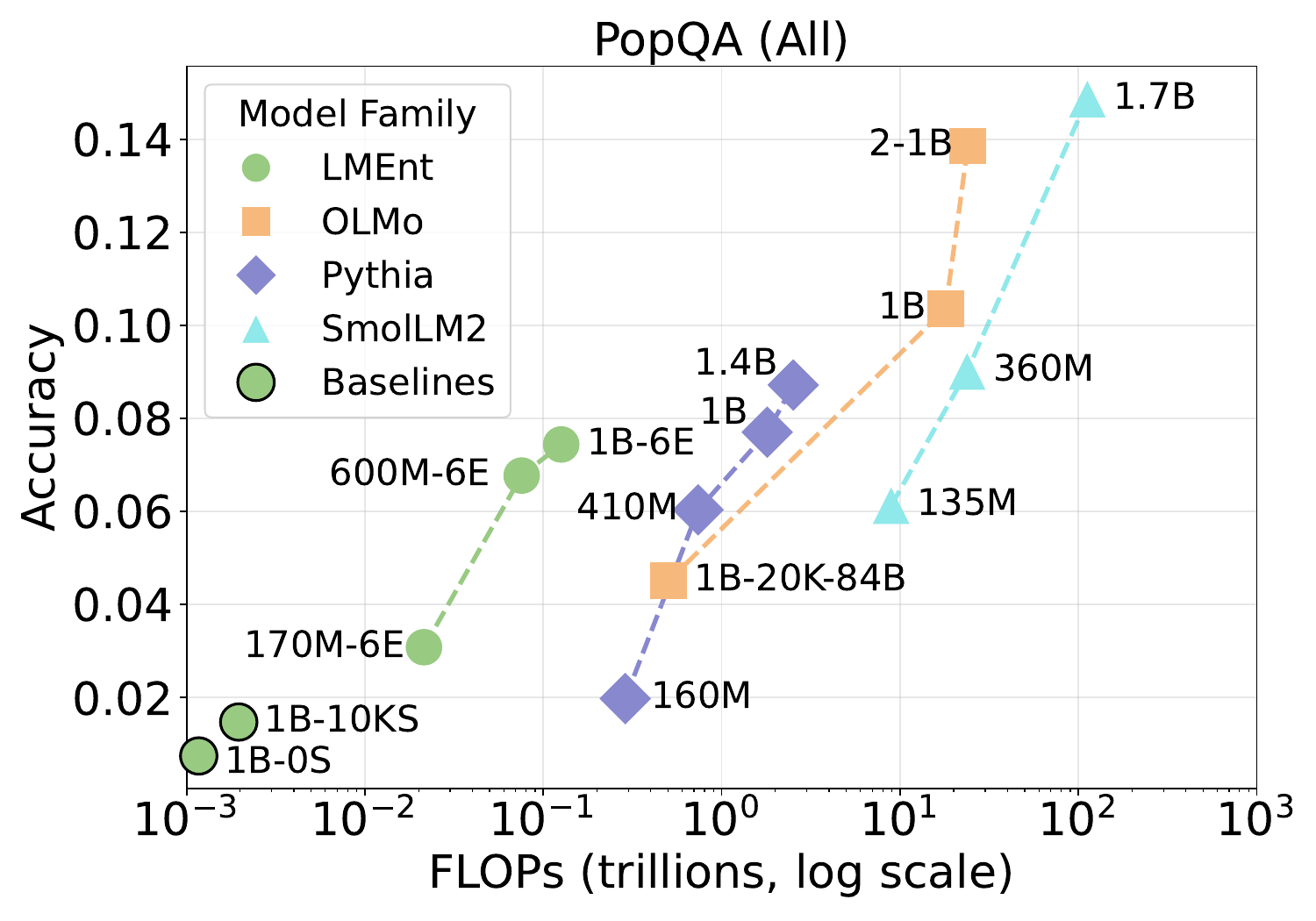}
    \includegraphics[width=.3\linewidth]{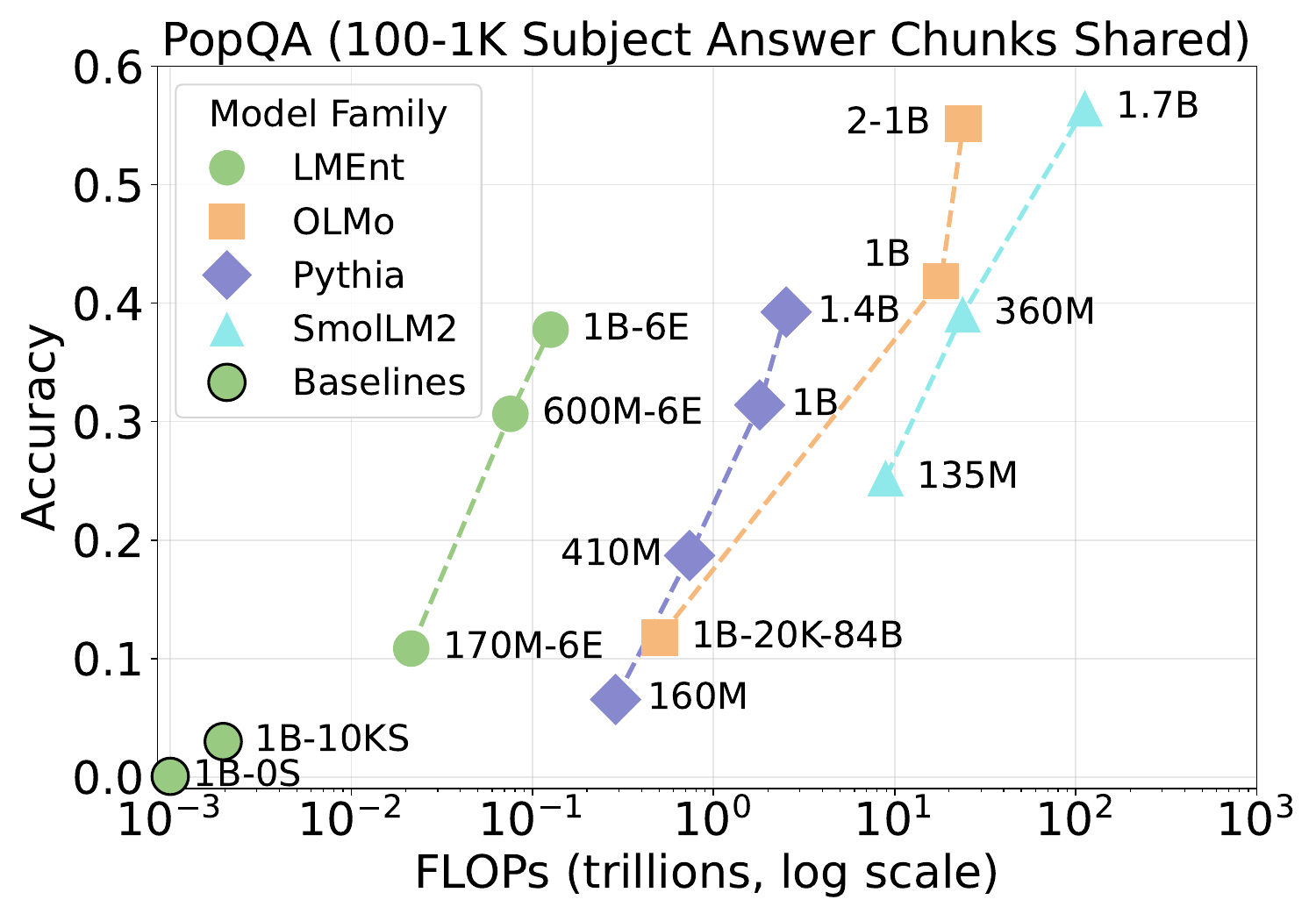}
    \caption{Accuracy on PopQA as a function of compute budget: (left) all questions and (right) questions for which the subject and answer entities appear together in 100-1K chunks (moderate).}
    \label{fig:popqa-flops-all}
\end{figure}

\begin{figure}[ht]
\setlength{\belowcaptionskip}{-15pt}
    \centering
    \includegraphics[width=.3\linewidth]{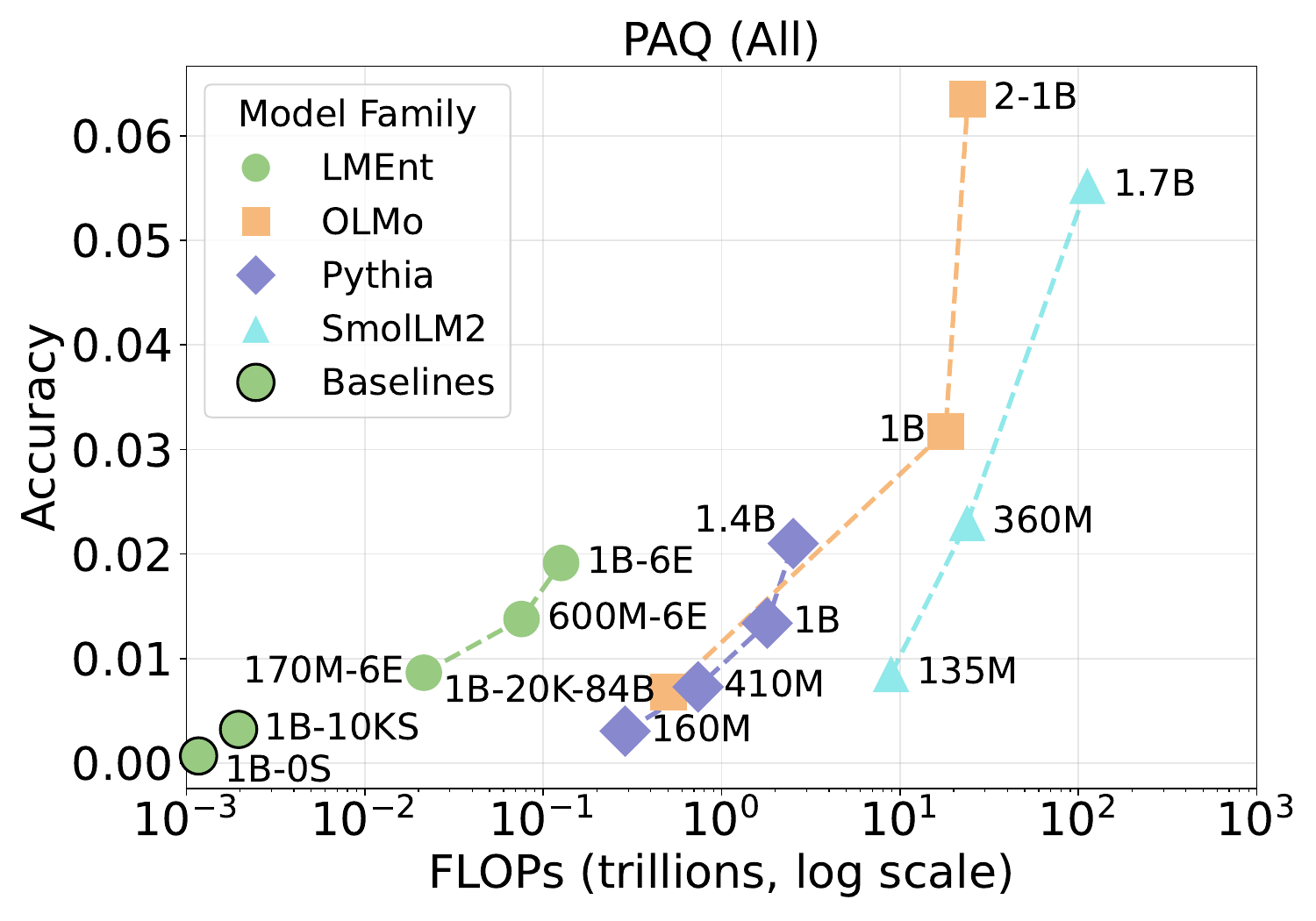}
    \includegraphics[width=.3\linewidth]{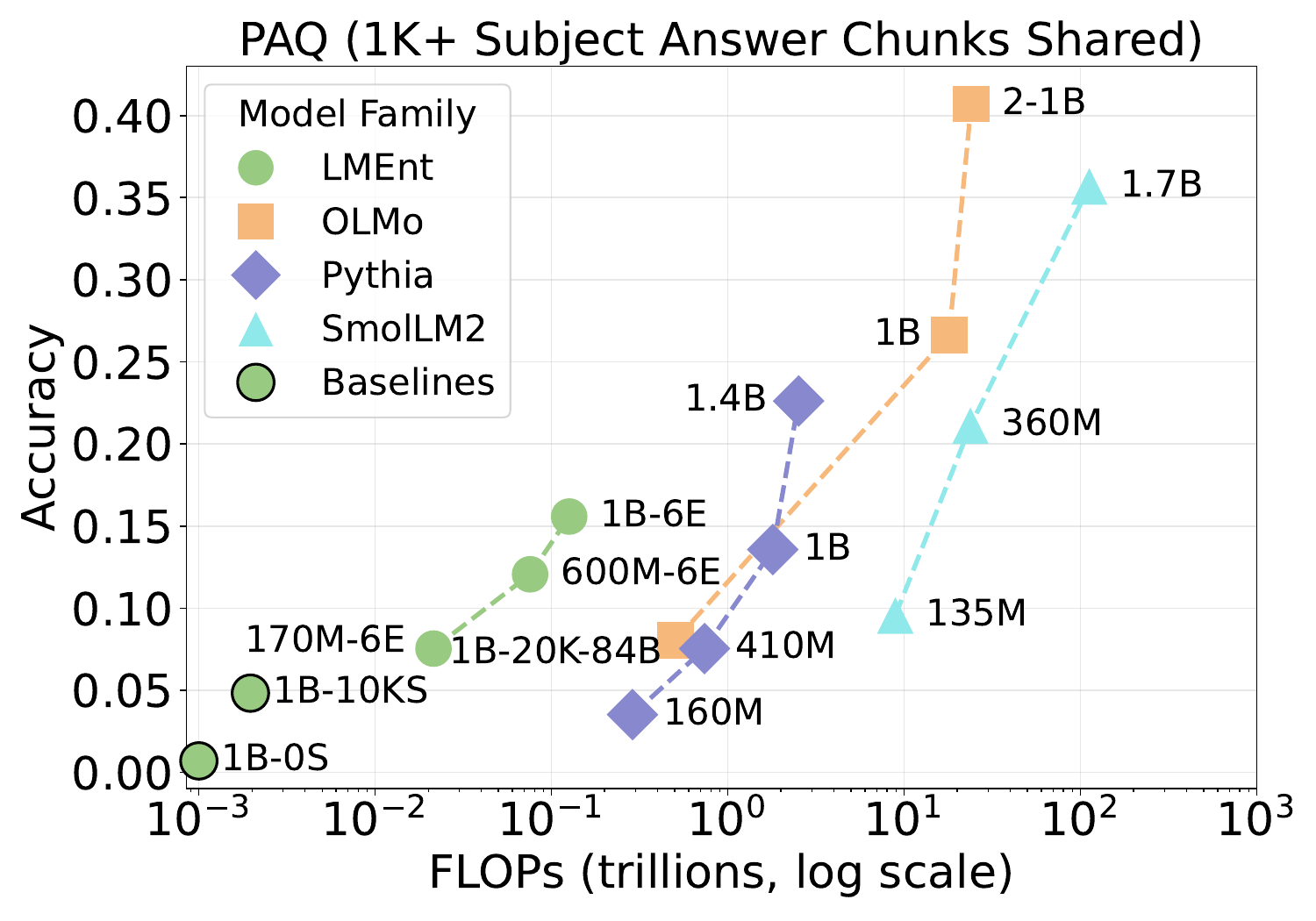}
    \caption{Accuracy on PAQ as a function of compute budget: (left) all entities and (right) questions for which the subject and answer entities appear together in 1K+ chunks (frequent).}
    \label{fig:paq-flops-all}
\end{figure}

In this section we present additional results for PopQA and PAQ accuracies as functions of compute budget. \Cref{fig:popqa-flops-all} shows the results for all questions and the subset of questions whose subject and answer entities co-occur moderately in PopQA, and \Cref{fig:paq-flops-all} shows the results for all questions and the subset of questions whose subject and answer entities co-occur frequently in PAQ. These results support \S\ref{sec:model-performance}.

\subsection{Popularity and Fact Frequency Indicators for Model Behavior on Additional Models}
\label{sec:other-model-indicators}
\begin{figure}[ht]
\setlength{\abovecaptionskip}{4pt}
\setlength{\belowcaptionskip}{-15pt}
    \centering
    \includegraphics[width=.32\linewidth]{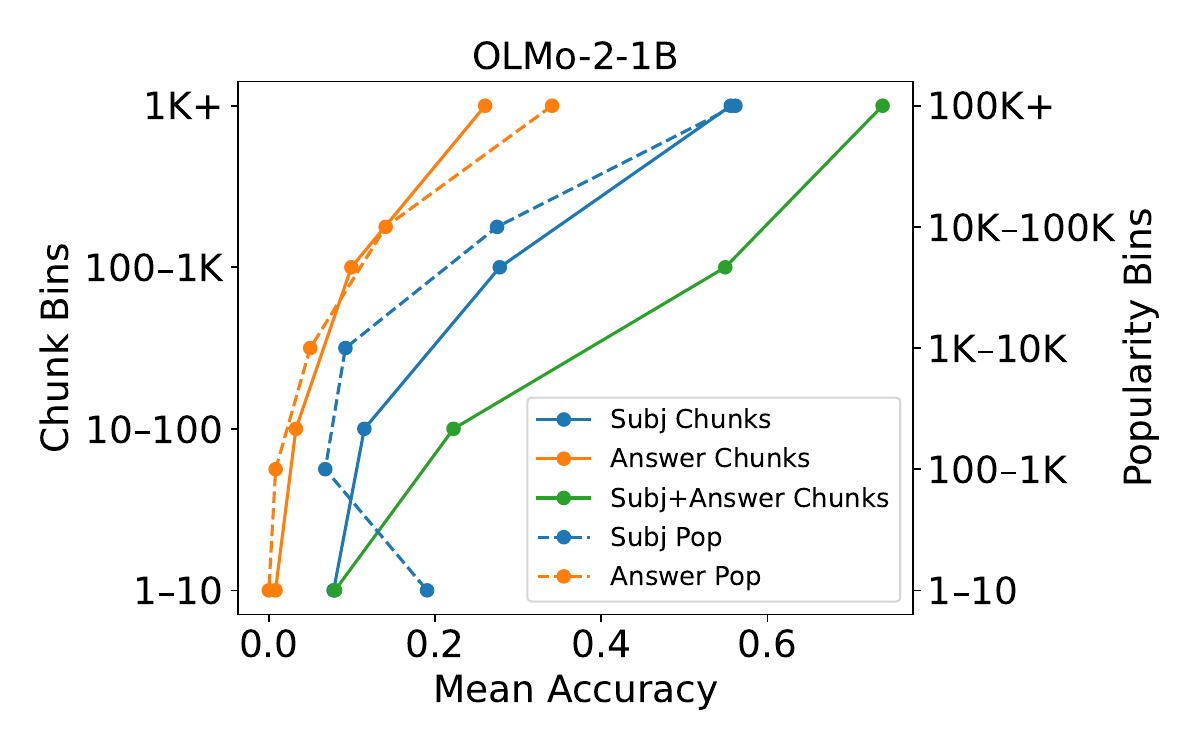}
    \includegraphics[width=.32\linewidth]{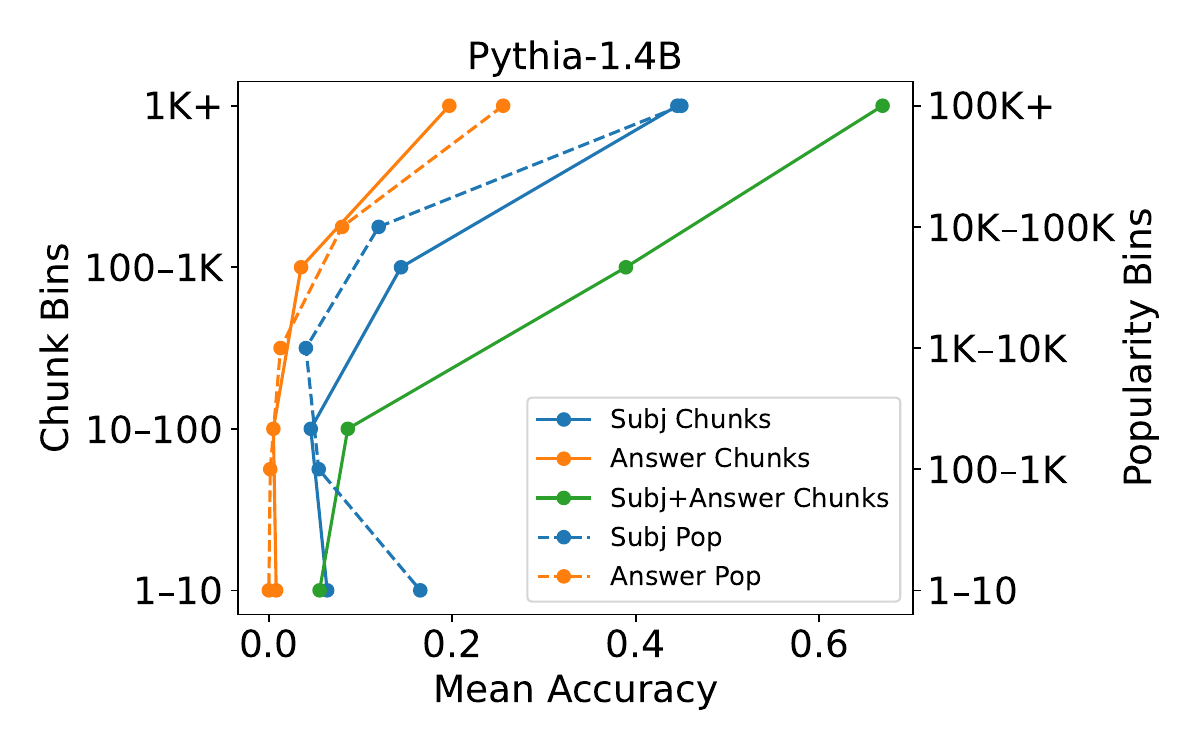}
    \includegraphics[width=.32\linewidth]{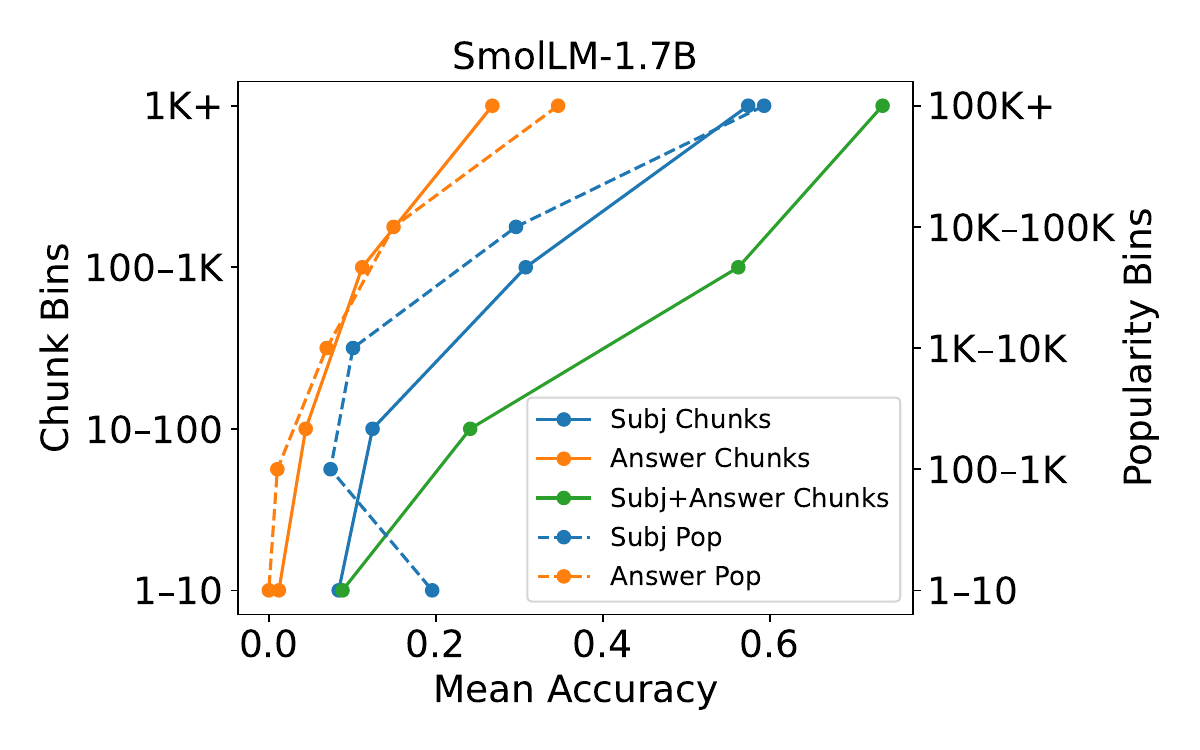}
    \caption{Accuracy of OLMo-2-1B \citep{olmo2}, Pythia-1.4B \citep{pythia}, and SmolLM-2-1.7B \citep{smollm} on PopQA, sliced by various indicators of popularity and fact frequency on PopQA.}
    \label{fig:other-model-indicators-all}
\end{figure}

In this section, we present additional results for PopQA accuracies sliced by popularity and fact frequency indicators. \Cref{fig:other-model-indicators-all} shows the results for OLMo-2-1B \citep{olmo2}, Pythia-1.4B \citep{pythia}, and SmolLM-2-1.7B \citep{smollm}. These results support \S\ref{sec:index-eval}.

\subsection{Results on Additional Tasks}
\label{sec:results-not-knowledge}
We report results on non-knowledge tasks in \Cref{tab:model-benchmark-results}, which complement \S\ref{sec:model-performance}.
\begin{table}[htbp]
\centering
\setlength{\belowcaptionskip}{5pt}
\caption{Performance of \method{} models and baseline models on commonsense reasoning, multiple choice, and reading comprehension tasks.}
\label{tab:model-benchmark-results}
\resizebox{\textwidth}{!}{%
\begin{tabular}{lcccccccccccccccccc}
\toprule
Model & arc\_challenge & arc\_easy & boolq & copa & coqa & drop & gsm8k & hellaswag & jeopardy & naturalqs\_open & openbookqa & piqa & sciq & squad & triviaqa & truthfulqa & winogrande \\
\midrule
LMEnt-170M-1E & 0.217 & 0.370 & 0.574 & 0.570 & 0.072 & 0.022 & 0.020 & 0.271 & 0.006 & 0.002 & 0.304 & 0.541 & 0.633 & 0.038 & 0.000 & 0.471 & 0.493 \\
LMEnt-170M-2E & 0.247 & 0.368 & 0.620 & 0.610 & 0.081 & 0.024 & 0.025 & 0.274 & 0.006 & 0.008 & 0.320 & 0.560 & 0.697 & 0.043 & 0.002 & 0.471 & 0.505 \\
LMEnt-170M-4E & 0.254 & 0.282 & 0.367 & 0.510 & 0.000 & 0.000 & 0.000 & 0.247 & 0.000 & 0.000 & 0.264 & 0.487 & 0.237 & 0.000 & 0.000 & 0.488 & 0.488 \\
LMEnt-170M-6E & 0.271 & 0.389 & 0.400 & 0.640 & 0.082 & 0.034 & 0.025 & 0.280 & 0.006 & 0.007 & 0.356 & 0.541 & 0.711 & 0.056 & 0.004 & 0.471 & 0.501 \\
\midrule
LMEnt-600M-1E & 0.271 & 0.409 & 0.604 & 0.650 & 0.085 & 0.012 & 0.025 & 0.285 & 0.008 & 0.001 & 0.340 & 0.555 & 0.722 & 0.044 & 0.001 & 0.472 & 0.490 \\
LMEnt-600M-2E & 0.301 & 0.405 & 0.620 & 0.630 & 0.113 & 0.046 & 0.055 & 0.296 & 0.009 & 0.020 & 0.350 & 0.554 & 0.744 & 0.085 & 0.005 & 0.468 & 0.516 \\
LMEnt-600M-4E & 0.254 & 0.389 & 0.565 & 0.610 & 0.150 & 0.055 & 0.015 & 0.306 & 0.034 & 0.010 & 0.338 & 0.559 & 0.735 & 0.092 & 0.016 & 0.470 & 0.515 \\
LMEnt-600M-6E & 0.274 & 0.409 & 0.625 & 0.630 & 0.167 & 0.075 & 0.050 & 0.318 & 0.023 & 0.029 & 0.374 & 0.569 & 0.762 & 0.119 & 0.045 & 0.444 & 0.510 \\
\midrule
LMEnt-1B-1E & 0.254 & 0.421 & 0.603 & 0.650 & 0.109 & 0.063 & 0.035 & 0.295 & 0.009 & 0.019 & 0.358 & 0.557 & 0.714 & 0.084 & 0.006 & 0.469 & 0.506 \\
LMEnt-1B-2E & 0.244 & 0.377 & 0.626 & 0.630 & 0.132 & 0.068 & 0.045 & 0.309 & 0.020 & 0.018 & 0.360 & 0.557 & 0.765 & 0.089 & 0.047 & 0.448 & 0.523 \\
LMEnt-1B-4E & 0.278 & 0.391 & 0.545 & 0.670 & 0.160 & 0.067 & 0.010 & 0.322 & 0.021 & 0.024 & 0.374 & 0.557 & 0.763 & 0.092 & 0.008 & 0.455 & 0.517 \\
LMEnt-1B-6E & 0.264 & 0.446 & 0.611 & 0.680 & 0.157 & 0.070 & 0.035 & 0.328 & 0.043 & 0.031 & 0.390 & 0.555 & 0.770 & 0.107 & 0.025 & 0.439 & 0.529 \\
\midrule
\midrule
OLMo-1B & 0.338 & 0.565 & 0.617 & 0.770 & 0.574 & 0.218 & 0.030 & 0.627 & 0.342 & 0.123 & 0.434 & 0.729 & 0.879 & 0.624 & 0.324 & 0.329 & 0.594 \\
OLMo-1B-20K-84B & 0.288 & 0.458 & 0.640 & 0.760 & 0.269 & 0.079 & 0.000 & 0.433 & 0.059 & 0.051 & 0.406 & 0.676 & 0.790 & 0.251 & 0.085 & 0.394 & 0.531 \\
OLMo-2-1B & 0.435 & 0.635 & 0.646 & 0.800 & 0.694 & 0.353 & 0.395 & 0.688 & 0.641 & 0.191 & 0.474 & 0.746 & 0.953 & 0.818 & 0.512 & 0.369 & 0.654 \\
\midrule
Pythia-1.4B & 0.348 & 0.539 & 0.582 & 0.760 & 0.566 & 0.206 & 0.020 & 0.540 & 0.273 & 0.098 & 0.434 & 0.707 & 0.874 & 0.565 & 0.250 & 0.386 & 0.566 \\
Pythia-1B & 0.331 & 0.525 & 0.623 & 0.740 & 0.519 & 0.185 & 0.025 & 0.478 & 0.195 & 0.077 & 0.378 & 0.683 & 0.890 & 0.534 & 0.188 & 0.389 & 0.533 \\
Pythia-410M & 0.301 & 0.467 & 0.588 & 0.700 & 0.428 & 0.166 & 0.040 & 0.393 & 0.069 & 0.058 & 0.360 & 0.663 & 0.833 & 0.438 & 0.104 & 0.410 & 0.532 \\
Pythia-160M & 0.311 & 0.423 & 0.498 & 0.640 & 0.218 & 0.136 & 0.020 & 0.300 & 0.019 & 0.031 & 0.324 & 0.606 & 0.746 & 0.176 & 0.042 & 0.442 & 0.502 \\
\midrule
SmolLM2-1.7B & 0.468 & 0.704 & 0.733 & 0.840 & 0.720 & 0.259 & 0.315 & 0.695 & 0.706 & 0.194 & 0.496 & 0.760 & 0.940 & 0.771 & 0.545 & 0.367 & 0.671 \\
SmolLM2-360M & 0.448 & 0.649 & 0.620 & 0.760 & 0.596 & 0.190 & 0.045 & 0.551 & 0.516 & 0.107 & 0.444 & 0.710 & 0.905 & 0.656 & 0.305 & 0.334 & 0.591 \\
SmolLM2-135M & 0.365 & 0.558 & 0.617 & 0.660 & 0.429 & 0.135 & 0.025 & 0.410 & 0.244 & 0.079 & 0.422 & 0.670 & 0.842 & 0.467 & 0.175 & 0.388 & 0.530 \\
\bottomrule
\end{tabular}
}
\end{table}

\subsection{Empirical Experiment for Choosing \method{} Score Thresholds}
\label{sec:score-thresholds-appendix}
In this section, we describe the empirical experiment that determined the score thresholds for \method{} retrieval (\S\ref{sec:chunk-retrieval-performance}). To determine thresholds for the hyperlink score $\textbf{H}$, entity-linking score $\textbf{EL}$, and coreference scores $\textbf{C}$ and $\textbf{CC}$, we sample 60 entities from PopQA based on hyperlinks counts. For each entity, we retrieve their scores using a fixed hyperlink threshold of $\textbf{H} = 1$, while varying $\textbf{EL}$, $\textbf{C}$, and $\textbf{CC}$ across the set $\{0.4, 0.5, 0.6, 0.7, 0.8\}$, and evaluate precision over chunks retrieved at different depths $k$. If $k > 100$, we randomly sample 100 of the $k$ returned chunks and submit them to Gemini 2.5 Flash Preview 6-17 using the prompt shown in \Cref{fig:gemini-prompt} to compute precision at $k$. Results are reported in \Cref{tab:dev-set-thresholds}.

\begin{table}[ht]
\setlength{\belowcaptionskip}{-5pt} 
\centering
\footnotesize
\caption{Empirical evaluation of \method{} retrieval precision across multiple thresholds and retrieval depths $k$ on the development set of 60 PopQA entities.}
\label{tab:dev-set-thresholds}
\begin{tabular}{lrrrrrrr}
\toprule
Score Thresholds & 1 & 5 & 10 & 100 & 1K & 10K & 100K \\
\midrule
H=1, EL = C = CC = 0.4 & 0.95 & 0.95 & 0.96 & 0.98 & 0.99 & 0.98 & 0.94 \\
H=1, EL = C = CC = 0.5 & 0.95 & 0.95 & 0.96 & 0.98 & 0.99 & 0.99 & 0.91 \\
H=1, EL = C = CC = 0.6 & 0.95 & 0.96 & 0.96 & 0.98 & 0.99 & 0.98 & 0.94 \\
H=1, EL = C = CC = 0.7 & 0.95 & 0.96 & 0.96 & 0.98 & 0.99 & 0.98 & 0.91 \\
H=1, EL = C = CC = 0.8 & 0.95 & 0.96 & 0.96 & 0.98 & 0.97 & 0.88 & 0.96 \\
\bottomrule
\end{tabular}
\end{table}

\subsection{Win Rates}
\label{sec:ss-win-rate-appendix}
\begin{figure}[ht]
\setlength{\abovecaptionskip}{4pt}
\setlength{\belowcaptionskip}{-20pt}
    \centering
    \includegraphics[width=.49\linewidth]{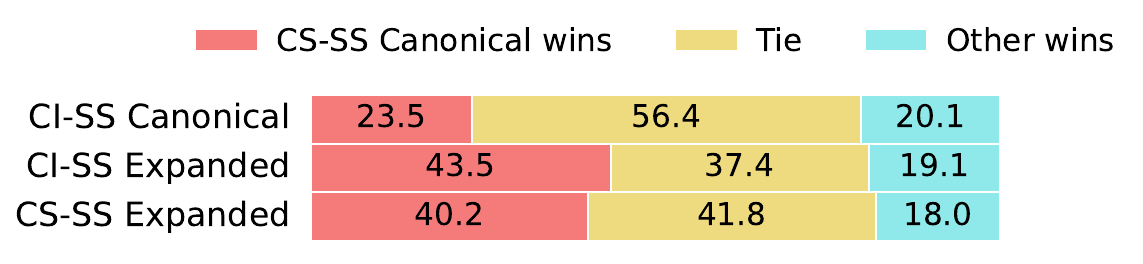}
   \includegraphics[width=.49\linewidth]{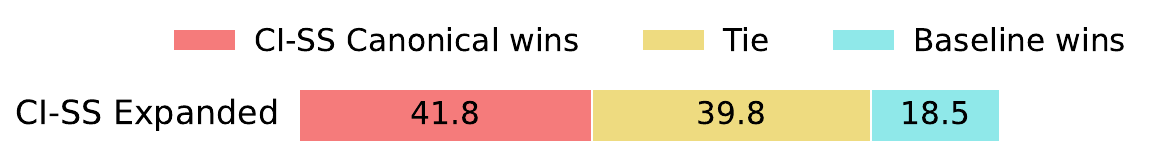}
    \caption{Pairwise win rates between \texttt{CS-SS Canonical} and other string-based methods (left). Pairwise win rates between \texttt{CI-SS Canonical} and \texttt{CI-SS Expanded} (right).}
    \label{fig:win-rate-summary-appendix}
\end{figure}

In \Cref{fig:win-rate-summary-appendix}, we extend the experiments in (\S\ref{sec:chunk-retrieval-performance}, \Cref{fig:win-rate-summary}) to compare win rates between the best string based method, \texttt{CS-SS Canonical}, and all other string-based variants (left), as well as \texttt{CI-SS Canonical} versus \texttt{CI-SS Expanded} (right). These results complement \Cref{fig:win-rate-summary}.

\subsection{\method{} provides better coverage for rare entities}
\label{sec:retrieval-rare}
\begin{figure}[t]
\setlength{\abovecaptionskip}{4pt}
\setlength{\belowcaptionskip}{-10pt}
    \centering
    \includegraphics[width=0.3\linewidth]{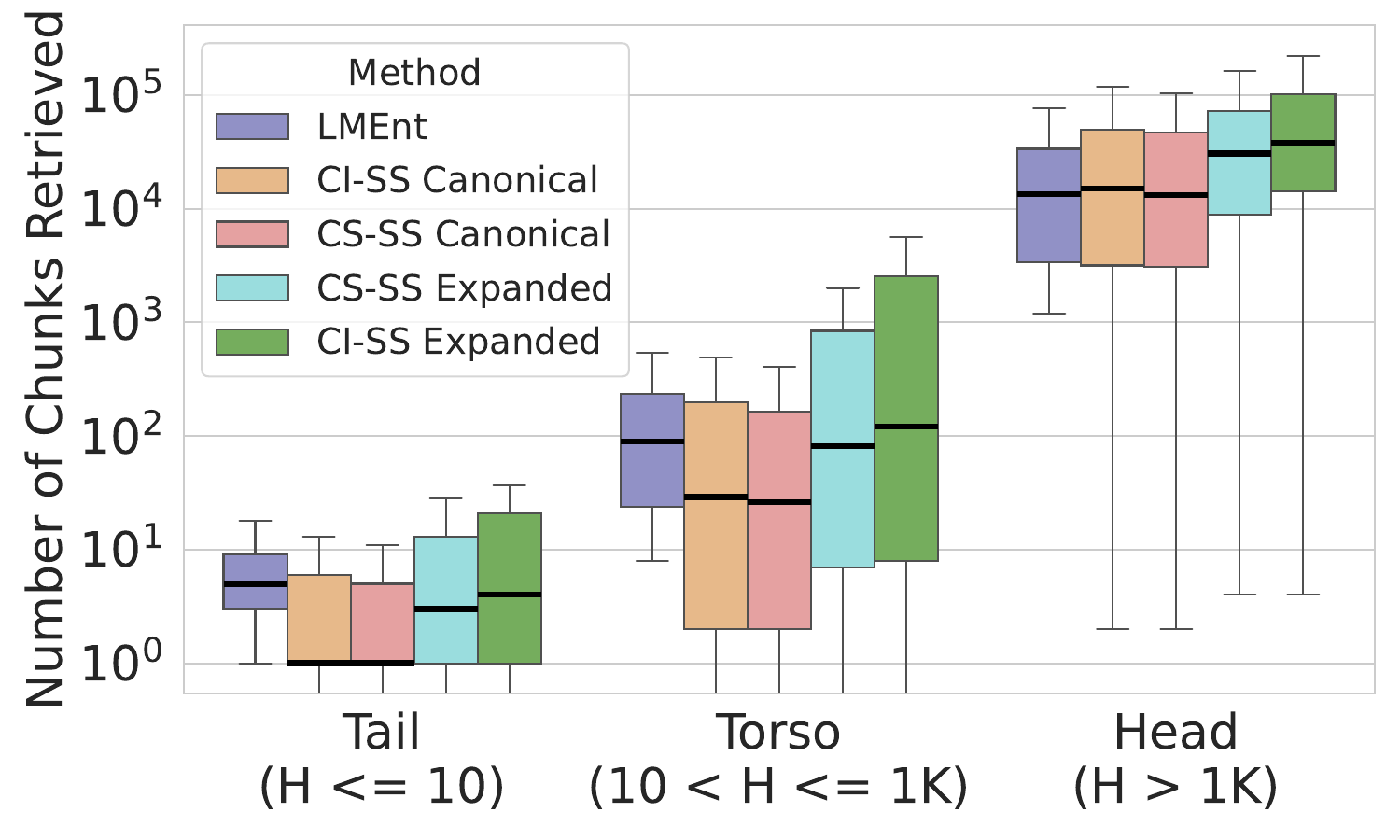}
    \caption{Distribution of the number of chunks retrieved by \method{}, and both \texttt{Canonical} and \texttt{Expanded} variants of \texttt{CI-SS} and \texttt{CS-SS}. Retrieval was performed for 1K PopQA entities, selected via stratified sampling based on hyperlink counts. \method{} retrieves more chunks than the \texttt{Canonical} variants for torso and tail entities, which together account for 99.7\% of all entities in Wikipedia. The higher number of chunks returned for head entities by \texttt{Expanded} variants is likely bloated by noisy retrieval (\Cref{fig:precision-at-k}).}
    \label{fig:retrieval-histogram}
\end{figure}

\Cref{fig:retrieval-histogram} shows the distribution over the total number of retrieved chunks per entity. We observe that \method{} offers greater coverage for tail and torso entities compared to \texttt{Canonical} string-based variants. This is especially significant since tail and torso entities together make up $99.7\%$ of the total entities in the corpus (\S\ref{sec:appendix-pretraining-data-stats}, \Cref{tab:datset_stats}). 
While it appears that \texttt{Expanded} string-based variants retrieve more chunks, this is likely noise due to alias ambiguity. There is also a noticeable difference in variance---specifically string-based methods show egregious retrieval failures, sometimes retrieving less than 10 documents for head entities and no documents for torso entities. These results support \S\ref{sec:index-eval}.

\subsection{\method{} wins across all entity frequency levels, often by large margins}
\label{sec:win-rate-detailed-appendix}
\begin{figure}[ht]
\setlength{\abovecaptionskip}{4pt}
\setlength{\belowcaptionskip}{-20pt}
    \centering
    \includegraphics[width=.7\linewidth]{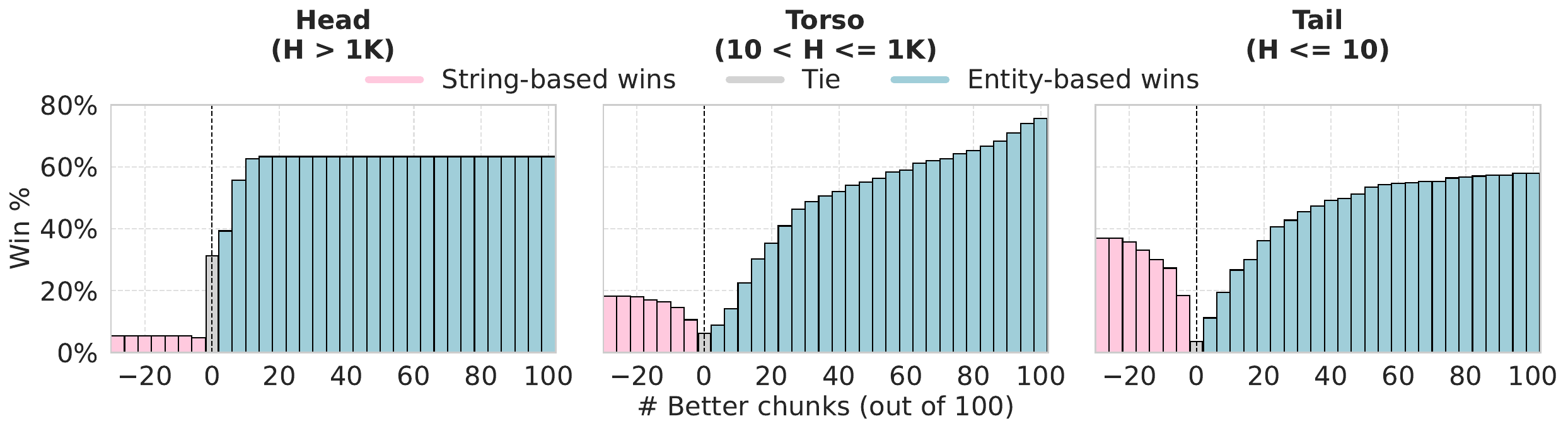}
    \caption{Comparison between \method{} retrieval and the best performing string-based approach (\texttt{CS-SS Canonical}). Results are grouped by frequency of hyperlinks; tail entities (right), torso entities (center), and head entities (left).}
    \label{fig:win-rate-distribution}
\end{figure}

\Cref{fig:win-rate-distribution} shows ``Yes'' chunk count differences between methods and the cumulative percentage of entities where one method outperforms the other, across entities of different frequencies. For tail entities (right), \method{} outperforms by a few chunks, but this is meaningful given the scarcity of mentions. For torso entities middle), \method{} wins by a substantial margin ($\geq 20$ additional correct chunks) in $40\%$ of cases. For head entities (left), the margin is typically smaller (1–25 chunks), suggesting that string-based search performs more competitively in these cases -- though, \method{} still outperforms it on $\geq 60\%$ of entities.

\subsection{Knowledge Acquisition during Pretraining}
\label{appendix:delta}

\begin{figure}[H]
\setlength{\belowcaptionskip}{-20pt}
    \centering
    \includegraphics[width=0.24\linewidth]{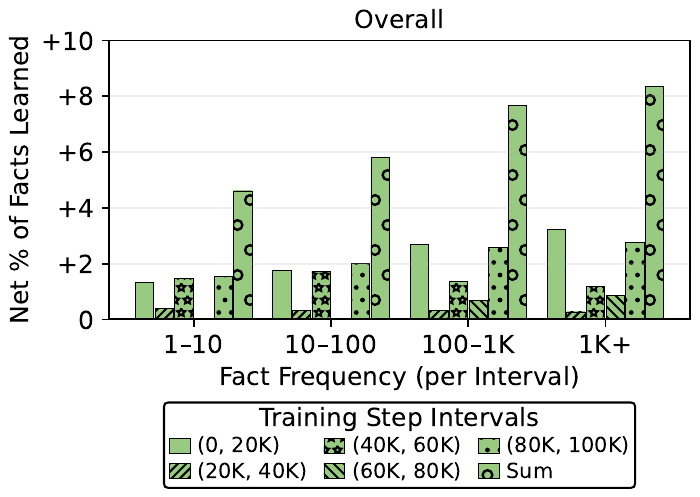}
    \includegraphics[width=0.24\linewidth]{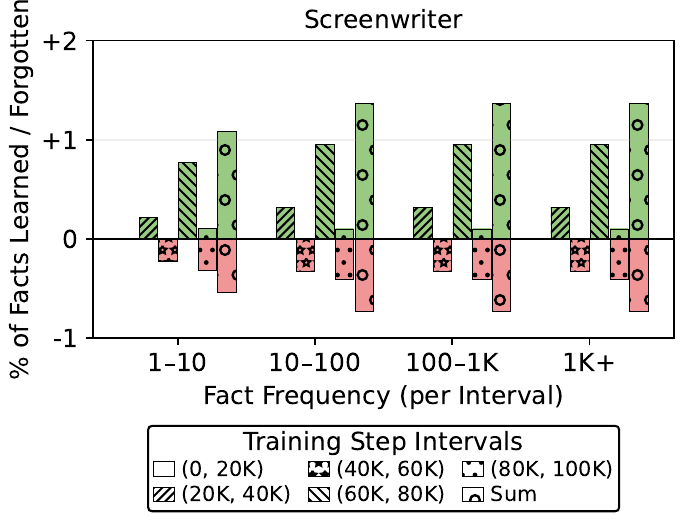}
    \includegraphics[width=0.24\linewidth]{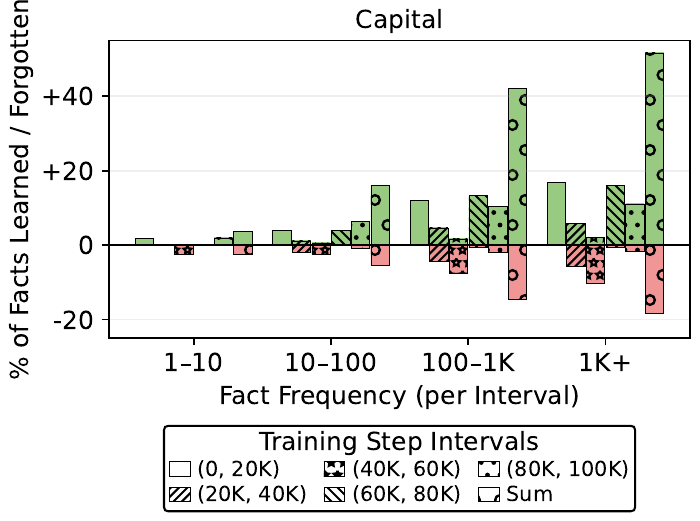}
    \includegraphics[width=0.24\linewidth]{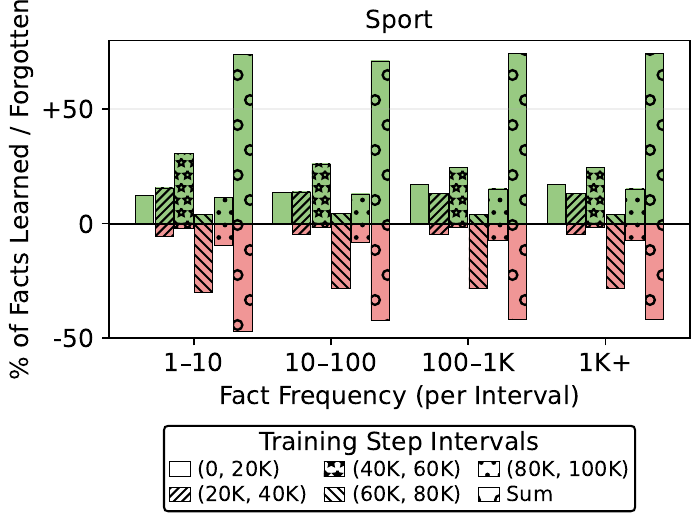}
    \caption{Percentage of net gains in facts learned between intermediate checkpoints of \method-1B-6E (left). Additional examples of the relation-specific analysis in in \S\ref{sec:plasticity}.}
    \label{fig:delta}
\end{figure}

\cref{fig:delta} shows the net gains in facts learned between \method-1B-6E checkpoints across frequency bins of subject and answer entity co-occurence,  and additional examples of three specific relations. The split of \% of facts learned and forgotten is in \S\ref{sec:plasticity}, \cref{fig:pos-mg}.

\newpage
\section{Appendix: Additional Implementation Details}

\subsection{Running Maverick At Scale}
\label{sec:run-maverick}
The backbone of Maverick \citep{maverick} is DeBERTa \citep{deberta} which has quadratic attention complexity with respect to sequence length. Therefore, running Maverick on long documents is impractical, e.g. 15K token long documents take 40-50GB VRAM. Therefore, we reduce the memory footprint, and accelerate and parallelize inference across multiple documents by replacing the DeBERTa backbone with a FlashAttention-based implementation \citep{FlashDeBERTa}. Then, we apply Maverick to non-overlapping windows of 50K tokens, which may result in multiple distinct coreference clusters for the same entity. In these cases, we rely on the coverage of hyperlinks and the entity linker (which runs on the entire document) to map each cluster to the target entity's QID.

\subsection{Coreference scoring}
\label{sec:coreference-scoring-appendix}
In this section we detail how we compute the score derived from coreference resolution, and how we resolve ambiguity when multiple entities are associated with a single coreference cluster (\S\ref{subsec:clustering-entity-mentions}).

\paragraph{Coref (C)} is designed to associate shortened references, not previously found by hyperlinks or entity linking, with their canonical names, e.g. \textcolor{lavender}{``\textit{the hospital}''} with \textcolor{lavender}{``\textit{John R. Oishei Children's Hospital''}}. The coreference cluster of \textcolor{lavender}{``\textit{the hospital}''} contains \textcolor{lavender}{``\textit{the hospital}''} and \textcolor{lavender}{``\textit{John R. Oishei Children's Hospital in}} \textcolor{peach}{Buffalo''}. There are two entities related to this cluster, with QIDs: \textcolor{lavender}{``\texttt{Q93565992}''} and \textcolor{darkpeach}{``\texttt{Q40435}''}. We can reduce ambiguity by leveraging the shared substring of ``\textit{(H,h)ospital}'' to promote \textcolor{lavender}{``\textit{the hospital}''} being mapped to 
\textcolor{lavender}{``\texttt{Q93565992}''} more strongly. We represent this promotion by a score denoted by \textbf{C}, which we formally define below.

Let the coreference cluster associated with $m$ be $\mathcal{C}$, and define the sets of hyperlink and entity linking mentions that overlap with $\mathcal{C}$ as $\mathcal{C_{\text{H}}}$ and $\mathcal{C_{\text{EL}}}$, respectively. To calculate $\textbf{C}(m, e)$, for each mention $m' \in \mathcal{C_{\text{H}}} \cup \mathcal{C_{\text{EL}}}$, we compute the longest common substring (LCS) with $m$. Let $\text{sim}(m, m')$ be the harmonic mean of the LCS overlap ratios between mention $m$ and $m'$, defined as:

\begin{equation}
\text{sim}(m, m') = 
\begin{cases}
0.0 & \text{if } \lvert m \rvert = 0 \text{ or } \lvert m' \rvert = 0 \\
\frac{2 \cdot a \cdot b}{a + b} & \text{otherwise}
\end{cases}
\quad \text{where } a = \frac{\text{LCS}(m, m')}{\lvert m \rvert},\;
b = \frac{\text{LCS}(m, m')}{\lvert m' \rvert}
\end{equation}

We weight either the hyperlink or entity linking score of $m'$ for $e$ by its textual similarity to $m$:

\begin{equation}
\text{support}(m, m', e) = \text{sim}(m, m') \cdot
\begin{cases}
\textbf{H}(m', e) & \text{if } m' \in \mathcal{C}_{\text{H}} \\
\textbf{EL}(m', e) & \text{if } m' \in \mathcal{C}_{\text{EL}}
\end{cases}
\end{equation}

Finally, the $\textbf{C}$ score for mention $m$ and entity $e$ is given by the maximum \text{support} score over all other mentions $m' \in \mathcal{C}$:
\begin{equation}
\textbf{C}(m, e) = \max_{m' \in \mathcal{C_{\text{H}}} \cup \mathcal{C_{\text{EL}}}} \text{support}(m, m', e)
\end{equation}

\paragraph{Coref-Cluster (CC)} connects generic indirect mentions (that don't share textual similarity with the entity name) to their canonical entity names, e.g. ``\textit{their}'' with ``\textit{Buffalo Bills}''. In this case, we compute a distribution of scores over all the entities linked to some mention in the cluster. The score $\textbf{CC}(e)$ is defined per entity per cluster, and is shared across all mentions $m \in \mathcal{C}$.

For each mention \(m \in \mathcal{C}\) and entity $e$, we look at the existing source scores and compute a $\text{m}_{\text{support}}$ score. We don't aggregate the existing source scores (e.g. compute an average), because we found that this disproportionately rewarded single mentions identified by multiple sources and outweighed multiple mentions found by single sources, leading to incorrect cluster to entity associations. The $\mathcal{C}_{\text{support}}$ for entity $e$ in cluster $\mathcal{C}$ is simply the sum over all $\text{m}_{\text{support}}$ scores.  

\begin{align}
\text{m}_{\text{support}}(m, e) &=
\begin{cases}
\textbf{H}(m, e) & \text{if } m \in \mathcal{C}_{\text{H}} \\
\textbf{EL}(m, e) & \text{if } m \in \mathcal{C}_{\text{EL}} \\
\textbf{C}(m, e) & \text{otherwise}
\end{cases}
\quad
\mathcal{C}_{\text{support}}(e) = \sum_{m \in \mathcal{C}} \text{m}_{\text{support}}(m, e)
\end{align}

To compute a distribution of scores over all entities mentioned in the cluster, we compute a softmax over $\textbf{CC}(\mathcal{C}, e')$ scores for the set of entities $e'$ supported by at least one mention in \(\mathcal{C}\), denoted by $\mathcal{E}_\mathcal{C}$.
\begin{equation}
\textbf{CC}(\mathcal{C}, e) = \frac{\exp\left(\mathcal{C}_{\text{support}}(e)\right)}{\sum_{e' \in \mathcal{E}_\mathcal{C}} \exp\left(\mathcal{C}_{\text{support}}(e')\right)}
\end{equation}

\subsection{Error Analysis of \method{} Annotations}
\label{sec:error-analysis-appendix}
\begin{figure}[ht]
    \centering
    \includegraphics[width=\linewidth]{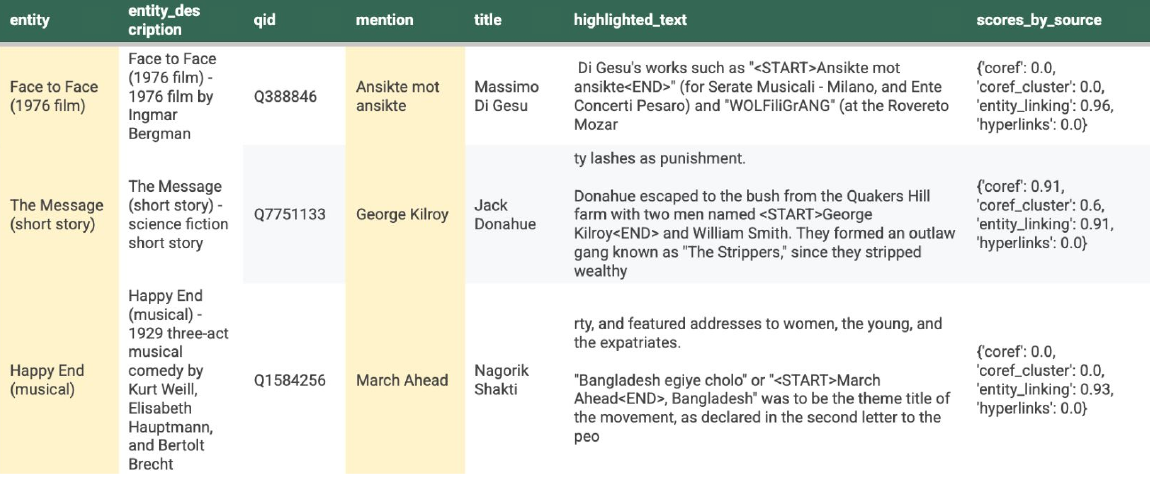}
    \caption{Error analysis of \method{} entity mentions.}
    \label{fig:error-analysis}
\end{figure}

This section describes a qualitative error analysis in which we sampled 112 mentions retrieved using \method{}, and manually analyzed the annotation errors observed and their frequency. We saw three errors total (2.7\%) that are displayed in \Cref{fig:error-analysis}. The errors were results of entity linking failures and one coreference resolution failure. 

\subsection{Pretrained Models}
\label{sec:models-appendix}

Details of the \method{} models are described here and support the overview in \S\ref{sec:overview}.
The 170M, 600M, and 1B models use $(\text{layers}, \text{hidden dimension})$ configurations of $(10, 768)$, $(16, 1344)$, and $(16, 2048)$, respectively. A hyperparameter search was conducted over the following ranges: global batch size $\{16{,}384,\ 32{,}768,\ 65{,}536,\ 131{,}072,\ 262{,}144\}$, peak learning rate $\{3 \times 10^{-4},\ 6 \times 10^{-4},\ 8 \times 10^{-4},\ 1.2 \times 10^{-3},\ 3 \times 10^{-3},\ 5 \times 10^{-3}\}$, and weight decay $\{0.005,\ 0.05,\ 0.1\}$. Hyperparameters were selected based on the minimal final training perplexity. All \method{} models were trained using the AdamW optimizer with a global batch size of $32{,}768$, rank batch size of $8{,}192$, peak learning rate of $5 \times 10^{-3}$, weight decay of $0.05$, and $1{,}000$ warmup steps.

\subsection{LLM Judge for Chunk Precision}
\label{sec:gemini-prompt-appendix}
To automatically evaluate the retrieval quality of both \method{} (our entity-based method) and string-based baselines (see \S~\ref{sec:chunk-retrieval-performance}), we use Gemini 2.5 Flash Preview 6-17. For each of the 1K entities in the test set, we apply \method{}, its ablations, and string-based methods \texttt{CS-SS-(Canonical, Expanded)} and \texttt{CI-SS-(Canonical, Expanded)}, to retrieve sets of relevant chunks. For a retrieved set containing more than 100 chunks, we randomly sample 100 chunks from it for evaluation.

\begin{figure}[ht]
    \centering
    \includegraphics[width=\linewidth]{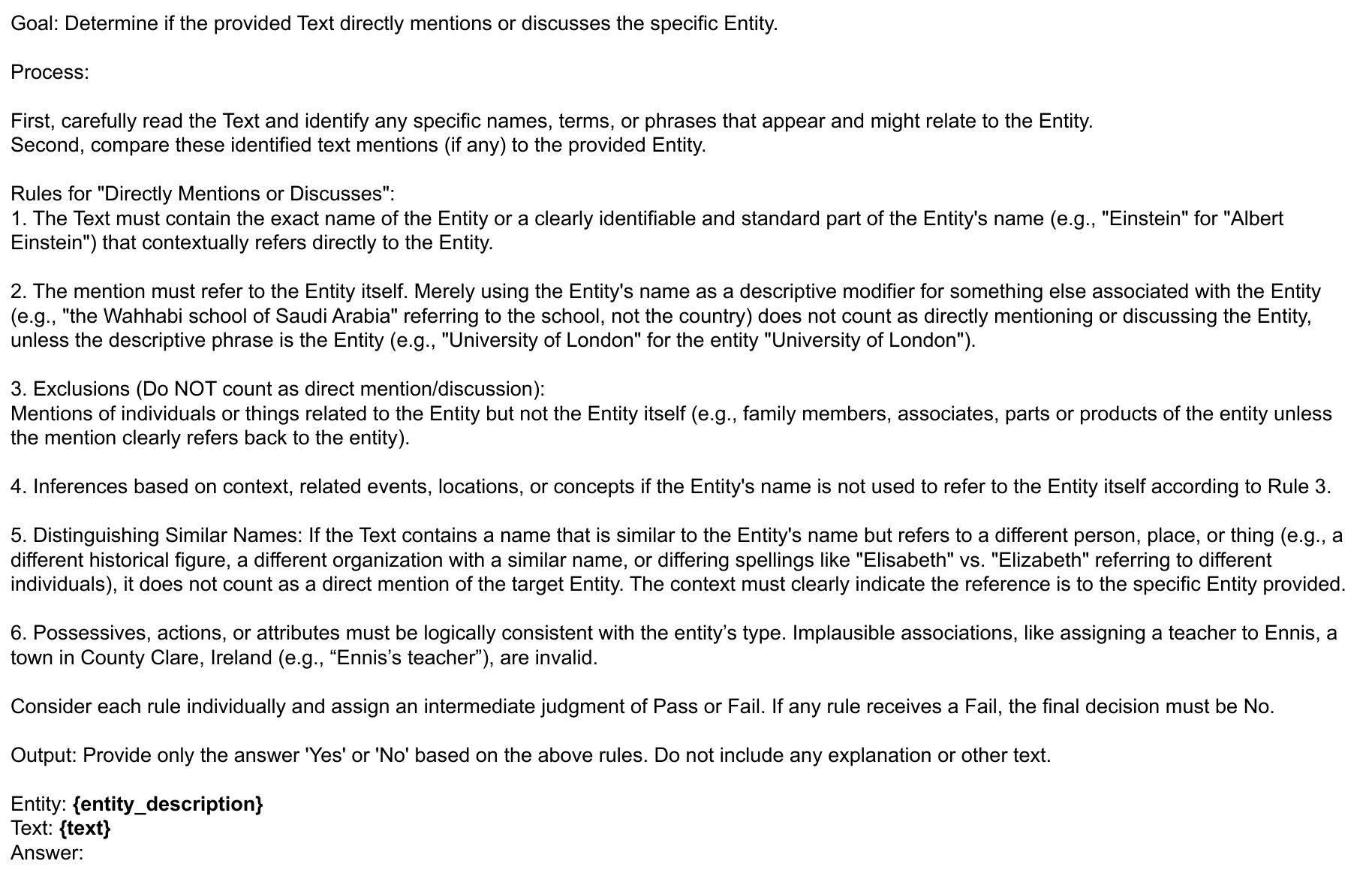}
    \caption{Prompt given to Gemini to automatically judge whether a chunk mentions an entity directly.}
    \label{fig:gemini-prompt}
\end{figure}

To determine whether a chunk mentions the entity, we prompt Gemini using the instruction shown in \Cref{fig:gemini-prompt}. To provide context, we include the entity's description from ReFinED \citep{refined}, which consists of the Wikipedia article title and the first sentence in the article. For each chunk, we identify matching mentions of the entity to create a short context window for Gemini. For \method{}, mentions are selected based on QID and score thresholds. For string-based methods, we use the \texttt{highlight} block in the Elasticsearch \citep{elasticsearch} query to extract character spans that match the entity name. Around each mention, we select a 130-character window to give Gemini some context of the mention in the chunk text. We evaluate up to three mentions per chunk. If Gemini returns ``Yes'' for any, the chunk is marked as mentioning the entity. If all are judged ``No'', the chunk is considered not to mention it. For \method{}, mentions are prioritized by a weighted sum of their scores, and for string-based methods, mentions are evaluated in the order they appear in the document.

\begin{figure}[ht]
    \centering
    \includegraphics[width=0.8\linewidth]{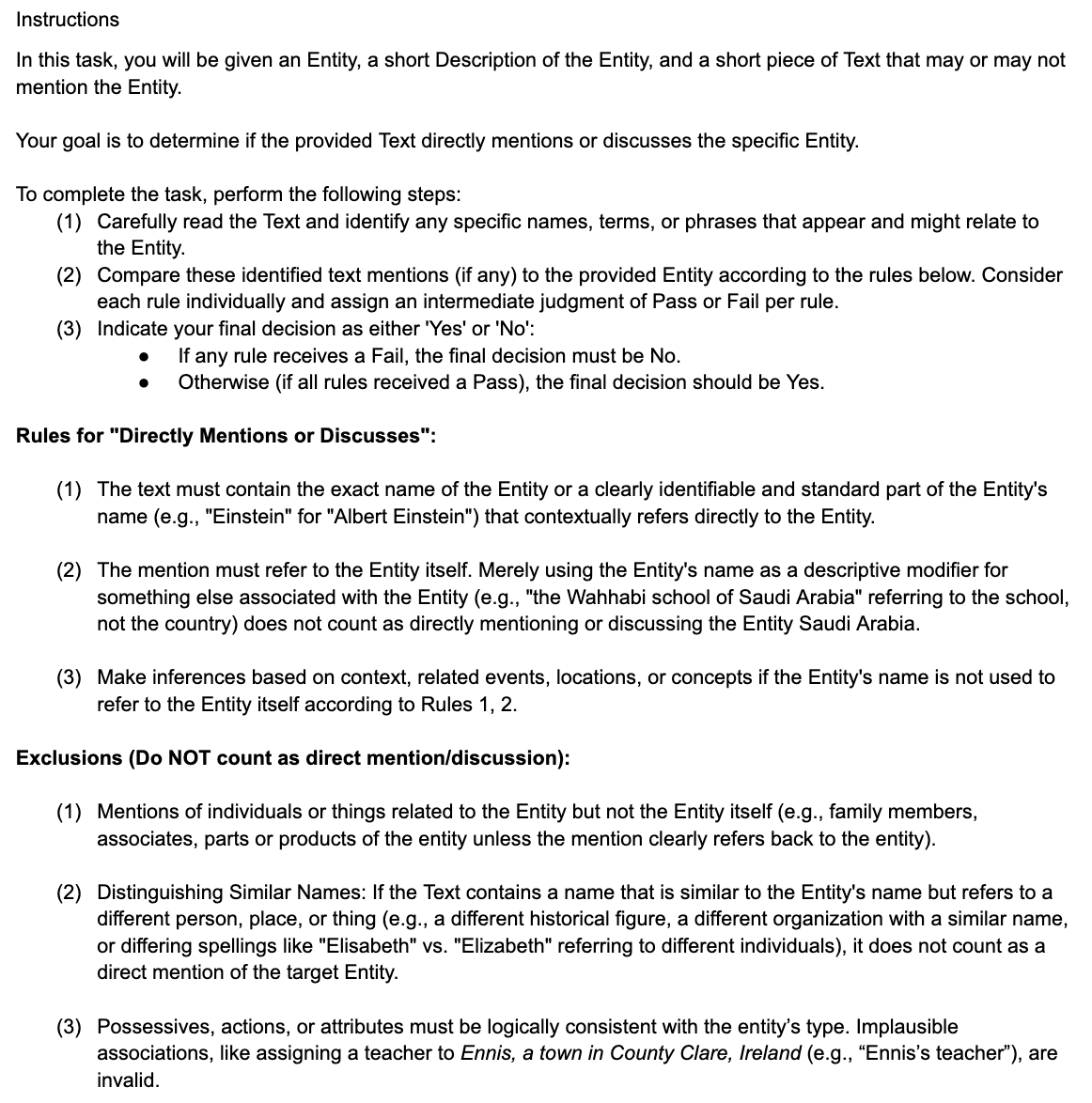}
    \caption{Instructions given to annotators for evaluating LLM-as-a-judge.}
    \label{fig:human-prompt}
\end{figure}

\paragraph{Justifying use of LLM-as-a-Judge}
To support our use of an LLM-as-a-Judge for evaluating model-generated responses, we use the alternative annotator test introduced by \citet{llm-judge-stat}. This test determines whether the LLM's performance is comparable or better than that of a randomly chosen human annotator. In line with their methodology, we employed three human annotators (graduate students) and used a dataset of 100 entity mentions randomly sampled from `Yes'' and ``No'' judged chunks. For each mention, the annotators received the same inputs as the LLM judge: the entity name, entity description, and 130-character context window around the mention, and were asked to evaluate whether the chunk named the entity. Instructions are found in \Cref{fig:human-prompt}. Following \citet{llm-judge-stat}, we set $\epsilon = 0.1$, which yielded a winning rate of $\omega=1.00$ with a $p$-values of $0.001, 0.001, \text{and } 8.28e^{-5}$ for all three annotators, indicating that the LLM can be relied on.

\subsection{Converting Queries to Cloze-Style Prompts}
\label{sec:paq-cloze-appendix}
\begin{figure}[ht]
    \centering
    \includegraphics[width=0.75\linewidth]{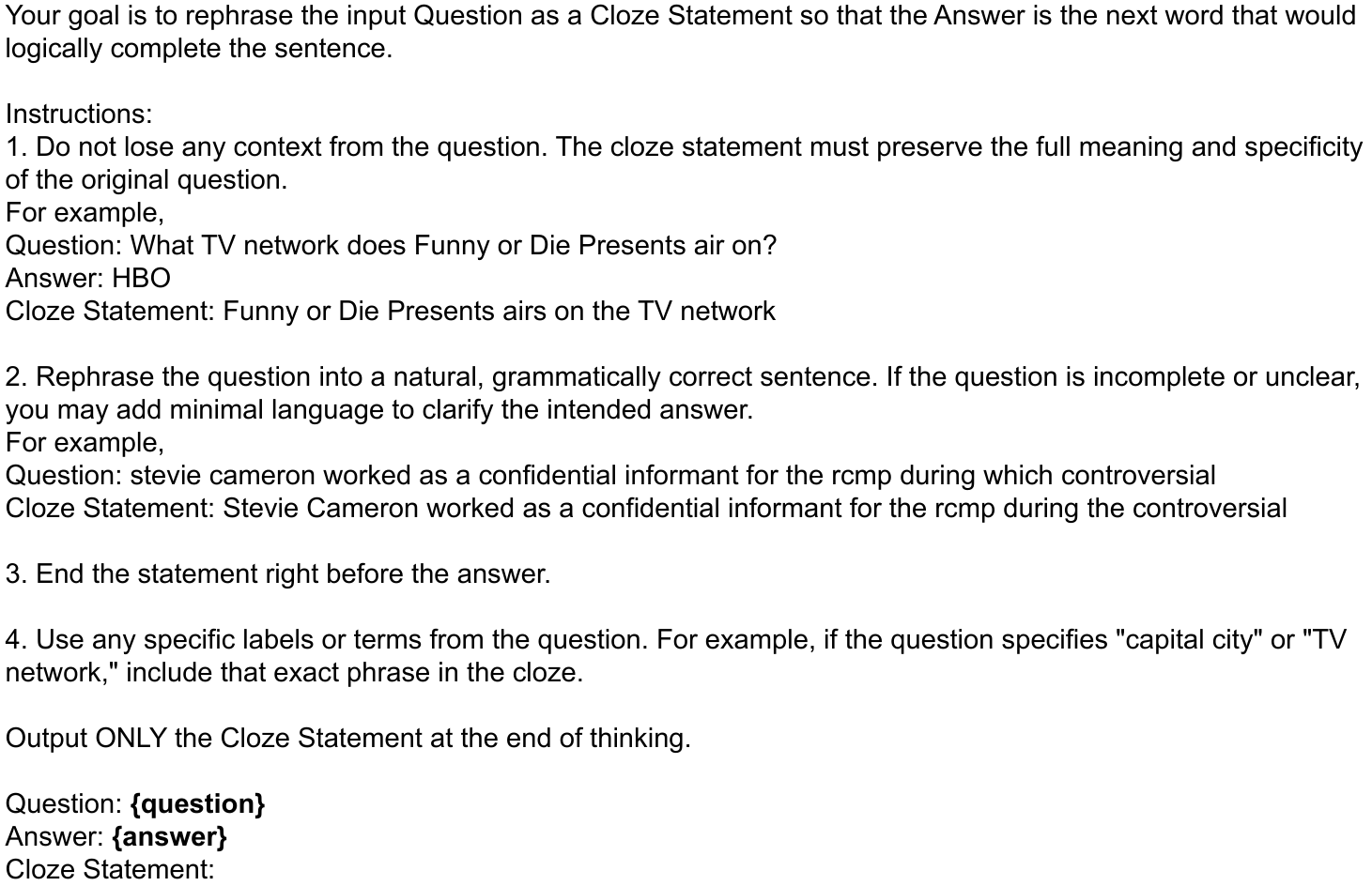}
    \caption{Prompt given to Gemini 2.5 Flash Preview 6-17 to convert PAQ questions to cloze-style prompts.}
    \label{fig:cloze-prompt}
\end{figure}

Since \method{} models are not instruction-tuned, we evaluate them on PAQ by converting each question into a cloze-style prompt, where the expected answer is the next predicted phrase. To perform this transformation, we use Gemini 2.5 Flash Preview 6-17 with the prompt shown in \Cref{fig:cloze-prompt}.

\end{document}